\documentclass[preprint]{article}

\usepackage[preprint]{neurips_2026}

\usepackage[utf8]{inputenc}
\usepackage[T1]{fontenc}
\usepackage{amsmath,amssymb,amsthm}
\usepackage{mathtools}
\usepackage{booktabs}
\usepackage{multirow}
\usepackage{graphicx}
\usepackage{xcolor}
\usepackage{hyperref}
\usepackage{cleveref}
\usepackage{algorithm}
\usepackage{algorithmic}
\usepackage{subcaption}
\graphicspath{{images/}}

\newtheorem{proposition}{Proposition}
\newtheorem{remark}{Remark}
\newtheorem{definition}{Definition}

\newcommand{\Wdist}{W_1}
\newcommand{\reals}{\mathbb{R}}

\DeclareMathOperator{\diag}{diag}

\DeclareMathOperator{\AUROC}{AUROC}
\DeclareMathOperator{\AUPRC}{AUPRC}
\DeclareMathOperator{\MAD}{MAD}

\title{Spectral Forensics of Diffusion Attention Graphs \\
       for Copy-Move Forgery Detection}

\author{%
  H.\,M.~Shadman Tabib$^{1}$ \quad
  Tasriad Ahmed Tias$^{1}$ \quad
  Nafis Tahmid$^{1}$ \\[3pt]
  $^{1}$Department of Computer Science and Engineering \\
  Bangladesh University of Engineering and Technology (BUET) \\
  Dhaka, Bangladesh
}

\begin{document}
\maketitle

\begin{abstract}
Copy-move forgery, where a region within an image is duplicated to conceal or fabricate content, remains a persistent threat to visual media integrity.
We introduce \textbf{GraphSpec\-Forge}, a training-free forensic framework that detects copy-move forgery by analysing the spectral properties of attention graphs extracted from a pretrained Stable Diffusion U-Net.
Our key theoretical insight is that copy-move manipulation induces \emph{approximate subgraph duplication} in the self-attention affinity graph, causing detectable spectral redistribution in the normalized graph Laplacian.
We formalise this connection via perturbation-theoretic arguments and construct an image-level anomaly detector based on Wasserstein transport distances between per-image Laplacian spectra and an authentic reference distribution.
We validate GraphSpecForge across \emph{four} copy-move benchmarks without forgery-specific retraining.
On the large-scale RecodAI-LUC benchmark (5{,}128 images), our best configuration achieves AUROC $= 0.606$ ($95\%$ CI: $0.580$--$0.638$; permutation $p = 0.005$), with the normalized Laplacian consistently outperforming raw attention spectra by $+0.057$ AUROC.
On three additional public benchmarks---MICC-F220, CoMoFoD, and COVERAGE---the same training-free pipeline attains AUROCs of $0.752$, $0.774$, and $0.673$, respectively; on CoMoFoD the detector further reaches AUPRC $= 0.833$, balanced accuracy $= 0.712$, and MCC $= 0.499$, with TPR$@1\%$FPR $= 32.5\%$.
The Laplacian advantage and the decoder-heavy layer concentration are reproduced on every benchmark, while the optimal feature bundle and fusion depth adapt to dataset scale---establishing a portable spectral anomaly mechanism rather than a brittle recipe.
Extended ablation and falsification experiments on a 400-image subset further confirm the signal's specificity and monotonic sensitivity to manipulation strength, and null-graph controls rule out trivial-statistic explanations.
We provide comprehensive ablations over spectral representations, feature groups, layer fusion strategies, and scoring methods, and present an honest assessment of the signal strength and remaining challenges for purely spectral forensic approaches.
\end{abstract}

\section{Introduction}
\label{sec:intro}

Copy-move forgery (CMF) is one of the most common and pernicious forms of image manipulation, in which a region of an image is duplicated and pasted elsewhere within the same image to conceal or fabricate visual content~\cite{fridrich2003detection,christlein2012evaluation}.
The duplicated regions may undergo geometric transformations, compression artefacts, or blending, making detection challenging.
Traditional CMF detectors rely on hand-crafted features such as block matching~\cite{popescu2004exposing}, keypoint descriptors~\cite{amerini2011sift}, or dense field methods~\cite{cozzolino2015efficient}.
More recent approaches leverage deep learning to learn discriminative representations~\cite{wu2018busternet,chen2021serial}, but these require task-specific training data and may not generalise across manipulation types.

A parallel line of research has explored \emph{foundation models as forensic tools}~\cite{wang2023dire,ojha2023towards}, recognising that large pretrained models encode rich structural priors about natural images.
Diffusion models~\cite{ho2020denoising,rombach2022high,uddin2025cryo}, in particular, build detailed internal representations through their denoising process, and their self-attention maps capture fine-grained spatial relationships.
Recent work has shown that diffusion attention maps can ground semantic concepts~\cite{tang2023daam,hertz2023prompt} and detect out-of-distribution manipulations~\cite{wang2023dire}.

In this paper, we pursue a fundamentally different approach: rather than using diffusion models as feature extractors for a downstream classifier, we analyse the \emph{graph-spectral structure} of their attention maps directly.
Our core hypothesis is:

\begin{quote}
\emph{Copy-move forgery induces approximate duplicated-subgraph perturbations in the self-attention affinity graph of a pretrained diffusion model, and these perturbations produce detectable spectral redistribution in the normalized graph Laplacian.}
\end{quote}

We formalise this hypothesis through perturbation-theoretic arguments (\Cref{sec:theory}), showing that subgraph duplication compresses local eigenvalue spacings and redistributes spectral mass.
We then construct \textbf{GraphSpecForge}, a fully training-free forensic pipeline (\Cref{sec:method}) that:
\begin{enumerate}
    \item Extracts self-attention affinity matrices from 16 layers of a pretrained Stable Diffusion v1.5 U-Net;
    \item Constructs normalized graph Laplacians and computes their eigenvalue spectra;
    \item Measures per-image spectral anomaly via Wasserstein transport distance to an authentic reference;
    \item Fuses scores across layers using reliability-weighted aggregation.
\end{enumerate}

\paragraph{Contributions.}
\begin{itemize}
    \item A \textbf{theoretical framework} connecting copy-move forgery to spectral perturbation of attention-derived graph Laplacians, grounded in classical results on graph duplication and eigenvalue interlacing.
    \item A \textbf{training-free image-level detector} achieving AUROC $0.606$ on the large-scale RecodAI-LUC benchmark ($n = 5{,}128$), with the normalized Laplacian consistently outperforming raw attention spectra ($+0.057$ AUROC, permutation $p = 0.005$).
    \item A \textbf{Forgery Spectral Expert Layer (FSEL)} selection methodology combining distributional, causal, and stability criteria to identify the most forensically informative attention layers, and a reproducible decoder-heavy ranking that transfers across datasets.
    \item \textbf{Multi-benchmark cross-dataset validation} on three additional public copy-move datasets (MICC-F220, CoMoFoD, COVERAGE) spanning JPEG compression, geometric transforms, and natural-scene duplications, reaching AUROCs of $0.752$, $0.774$, and $0.673$ and MCC up to $0.499$ \emph{without any forgery-specific retraining}, while a controlled Laplacian-vs.-raw comparison at matched configuration reveals dataset-dependent but consistently non-negative Laplacian advantage on the two larger-manipulation benchmarks.
    \item \textbf{Comprehensive ablations and falsification tests} over spectral representations, feature groups, layer fusion strategies, normalisation methods, null-graph controls, non-copy-move corruption negatives, and a controlled severity sweep---giving an honest multi-dataset assessment of signal strength and remaining limitations.
\end{itemize}

\begin{figure}[t]
\centering
\includegraphics[width=\linewidth]{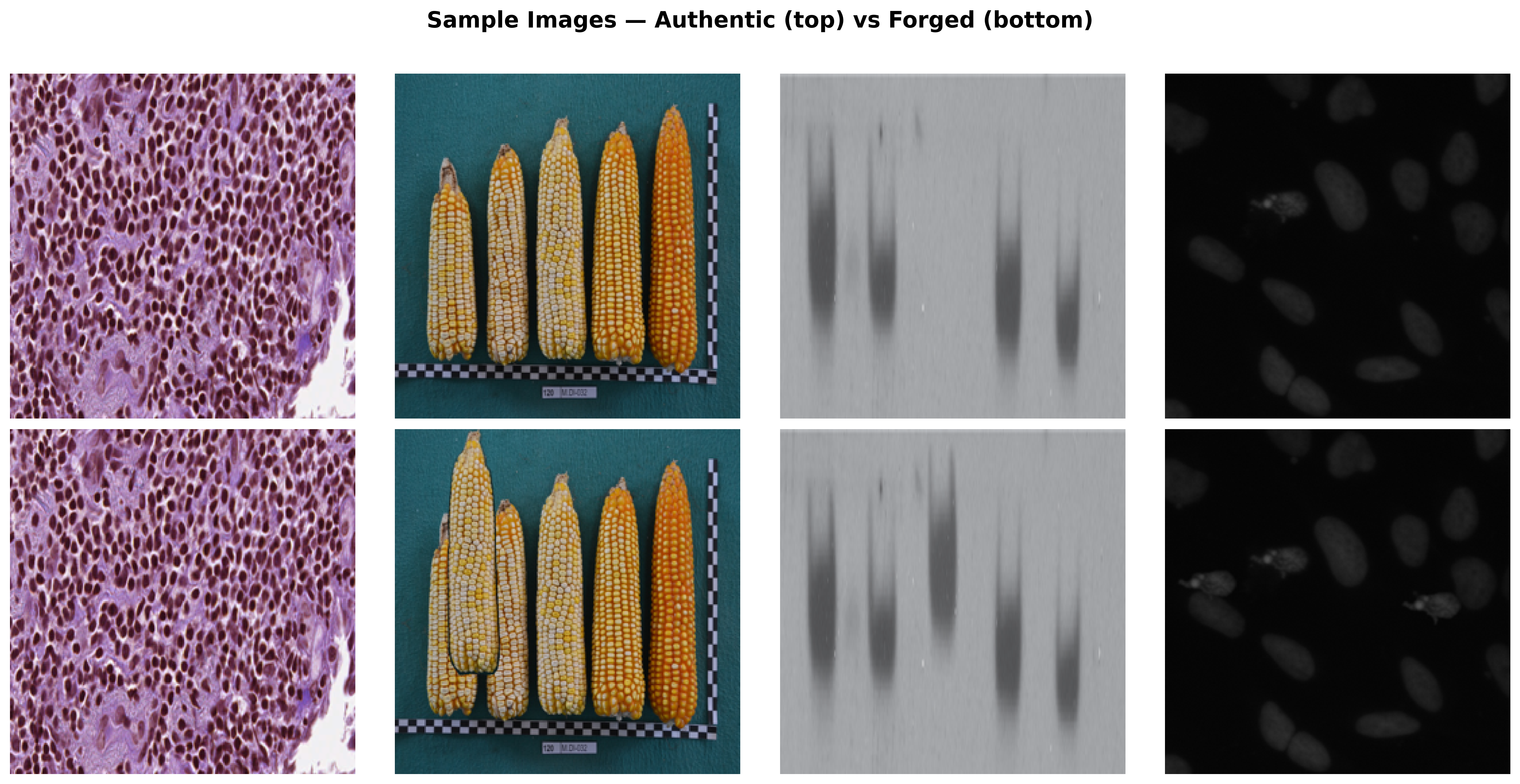}
\caption{\textbf{Sample images from the RecodAI-LUC dataset.} Top row: authentic images. Bottom row: corresponding forged copies with copy-move manipulations. The duplicated regions are visually subtle, making detection challenging even for human observers. For anonymity and to avoid cluttering the figure, only the dataset identifier is stated here; individual source filenames are omitted.}
\label{fig:sample_images}
\end{figure}

\section{Related Work}
\label{sec:related}

\paragraph{Copy-move forgery detection.}
Classical methods detect duplicated regions through block matching with DCT~\cite{fridrich2003detection}, PCA~\cite{popescu2004exposing}, or Zernike moments~\cite{ryu2010detection}, followed by keypoint-based approaches using SIFT~\cite{amerini2011sift} or SURF features.
Dense-field methods~\cite{cozzolino2015efficient} estimate pixel-level correspondence maps.
Deep learning approaches include BusterNet~\cite{wu2018busternet}, which uses a two-branch architecture, and DOA-GAN~\cite{islam2020doa}, which synthesises training data.
These methods typically require task-specific training and may struggle with unseen manipulation parameters.

\paragraph{Diffusion models for image forensics.}
Wang et al.~\cite{wang2023dire} proposed DIRE, which reconstructs images through a diffusion process and uses reconstruction error for detection.
Ojha et al.~\cite{ojha2023towards} demonstrated that features from large pretrained models generalise across manipulation types.
Corvi et al.~\cite{corvi2023detection} studied artefacts specific to diffusion-generated images.
Our approach differs fundamentally: we do not use the diffusion model as a feature extractor for classification, but analyse the structural properties of its internal attention representations.

\paragraph{Spectral methods in graph analysis.}
The normalized graph Laplacian $L = I - D^{-1/2}AD^{-1/2}$ is central to spectral graph theory~\cite{chung1997spectral}.
Its eigenvalues encode connectivity, clustering structure, and symmetries~\cite{von2007tutorial}.
Perturbation theory for graph spectra~\cite{stewart1990matrix} provides bounds on eigenvalue shifts under structural modifications.
The Weyl interlacing theorem constrains how eigenvalues change when vertices or edges are added~\cite{haemers1995interlacing}.
We leverage these classical results to predict how copy-move duplication should perturb attention graph spectra.

\paragraph{Attention map analysis.}
Self-attention mechanisms in vision transformers and diffusion models create implicit graphs over spatial tokens.
DAAM~\cite{tang2023daam} and related methods~\cite{hertz2023prompt} extract and interpret attention for semantic grounding.
The Seg4Diff framework~\cite{seg4diff2025} identifies ``expert layers'' in diffusion architectures whose attention maps best ground semantic regions.
We adapt this expert-layer concept to forensics, defining the \emph{Forgery Spectral Expert Layer} (FSEL) as the layer most sensitive to manipulation-induced spectral perturbation.

\section{Theoretical Framework}
\label{sec:theory}

\subsection{Attention as a Weighted Graph}
\label{sec:theory:graph}

Consider a self-attention layer $\ell$ in a diffusion U-Net processing image $x$.
The attention operation produces a matrix $A^{(\ell)} \in \reals^{n \times n}$, where $n$ is the number of spatial tokens.
Each entry $A^{(\ell)}_{ij}$ encodes the affinity between tokens $i$ and $j$.
We symmetrise this to obtain an undirected weighted graph:
\begin{equation}
    \bar{A}^{(\ell)} = \frac{1}{2}\bigl(A^{(\ell)} + {A^{(\ell)}}^\top\bigr), \quad
    \bar{A}^{(\ell)}_{ij} \leftarrow \max\!\bigl(\bar{A}^{(\ell)}_{ij},\, 0\bigr).
    \label{eq:symmetrise}
\end{equation}
The degree matrix $D = \diag(d_1, \ldots, d_n)$ with $d_i = \sum_j \bar{A}_{ij}$ yields the \emph{normalized graph Laplacian}:
\begin{equation}
    L = I - D^{-1/2}\,\bar{A}\,D^{-1/2},
    \label{eq:laplacian}
\end{equation}
whose eigenvalues $0 = \lambda_1 \le \lambda_2 \le \cdots \le \lambda_n \le 2$ encode the graph's structural properties.

\subsection{The Duplicated-Subgraph Hypothesis}
\label{sec:theory:hypothesis}

\begin{figure}[t]
\centering
\includegraphics[width=\linewidth]{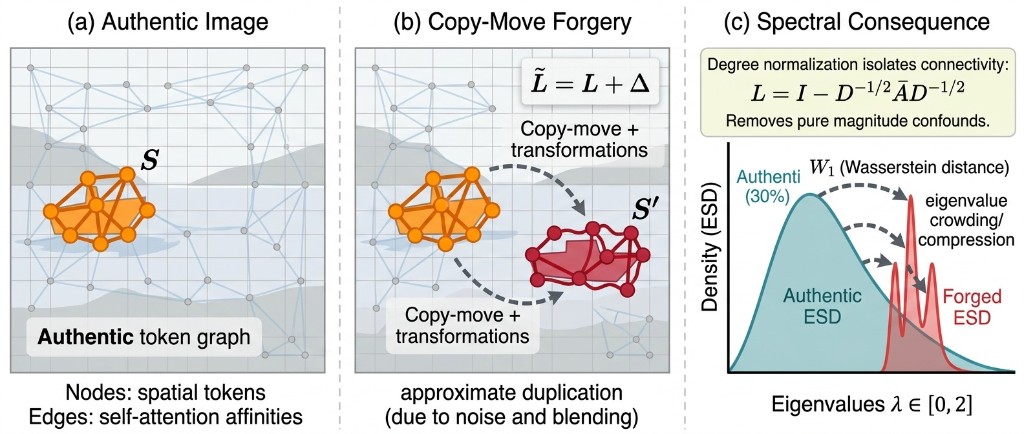}
\caption{\textbf{The duplicated-subgraph hypothesis.}
\textbf{(a)}~An authentic image's self-attention induces a token graph with subgraph~$S$.
\textbf{(b)}~Copy-move forgery approximately duplicates~$S$ to create~$S'$, perturbing the Laplacian $\tilde{L} = L + \Delta$ under real-world transforms and blending.
\textbf{(c)}~\emph{Representative illustration} of the resulting spectral redistribution: eigenvalue crowding/compression and a shift in the empirical spectral density, quantified by the Wasserstein transport distance $W_1$. Degree normalisation isolates connectivity structure from raw attention magnitude. Panel~(c) is schematic; quantitative per-layer ESDs from the real RecodAI-LUC benchmark are reported in \Cref{sec:results:layers} and \Cref{fig:pairwise_013}.}
\label{fig:theory_hypothesis}
\end{figure}

\Cref{fig:theory_hypothesis} illustrates the core intuition.
An authentic image's self-attention defines a weighted token graph~$G$ with a salient subgraph~$S$.
Copy-move forgery approximately duplicates~$S$ at a new location~$S'$, perturbing the Laplacian $\tilde{L} = L + \Delta$ and causing eigenvalue compression and spectral mass redistribution---detectable via the Wasserstein distance between authentic and forged \emph{empirical spectral distributions} (ESDs). Throughout this paper, the ESD of an $n \times n$ symmetric matrix $M$ with eigenvalues $\{\mu_k\}_{k=1}^n$ denotes the discrete probability measure $\mu_M = \frac{1}{n}\sum_{k=1}^{n}\delta_{\mu_k}$ that places mass $1/n$ on each eigenvalue; all transport distances and density-level comparisons in this paper are taken between such measures.

\begin{definition}[Approximate subgraph duplication]
Let $G = (V, E, w)$ be a weighted graph derived from attention on an authentic image.
Let $S$ be a subgraph of $G$ with vertex set $V(S) \subseteq V$. A \emph{copy-move perturbation} produces graph $\tilde{G}$ by approximately duplicating $S$: there exists a subgraph $S'$ of $\tilde{G}$ with vertex set $V(S') \subseteq V$ satisfying $|V(S')| \approx |V(S)|$, such that the induced subgraphs $G[S]$ and $\tilde{G}[S']$ have similar weighted edge structure, i.e., $\|w_S - \tilde{w}_{S'}\|_F \le \epsilon$ for small $\epsilon > 0$. The bound tolerates the geometric transforms, compression artefacts, and blending that accompany real-world copy-move operations, so the duplication is treated as approximate rather than exact.
\end{definition}

\begin{proposition}[Spectral redistribution under subgraph duplication]
\label{prop:spectral}
Let $L$ and $\tilde{L}$ be the normalized Laplacians of $G$ and $\tilde{G}$ respectively, where $\tilde{G}$ contains an approximate duplication of subgraph $S$.
Then:
\begin{enumerate}
    \item[(i)] \textbf{Eigenvalue compression}: If $S$ has $k$ connected components in the original graph, the duplication introduces at most $k$ additional near-degenerate eigenvalues, compressing local eigenvalue spacings.
    \item[(ii)] \textbf{Spectral mass redistribution}: The empirical spectral distribution $\mu_{\tilde{L}}$ satisfies
    $\Wdist(\mu_L, \mu_{\tilde{L}}) \le C \cdot \|L - \tilde{L}\|_F / n$,
    where $C$ depends on the spectral radius.
    \item[(iii)] \textbf{Band-specific shifts}: The perturbation concentrates spectral mass changes in frequency bands corresponding to the scale of the duplicated region.
\end{enumerate}
\end{proposition}

\begin{remark}
Part (i) follows from Weyl's interlacing inequality applied to the augmented graph.
Part (ii) is a consequence of the Hoffman--Wielandt inequality for symmetric matrices.
Part (iii) is an empirical observation consistent with the theory of graph wavelets~\cite{hammond2011wavelets}, where spatial scale maps to spectral frequency.
We emphasise that copy-move forgery in natural images produces \emph{approximate}, not exact, duplication due to geometric transforms, compression, and blending.
The perturbation framework $\tilde{L} = L + \Delta$ with $\|\Delta\|$ bounded accommodates these deviations.
\end{remark}

\subsection{Why Normalized Laplacian Outperforms Raw Attention Spectra}
\label{sec:theory:why_laplacian}

The raw attention matrix $A$ conflates two sources of variation: the overall attention magnitude (related to image content and style) and the structural connectivity pattern (related to spatial relationships).
The degree normalisation in \Cref{eq:laplacian} removes the magnitude confound, isolating the connectivity structure.
Under copy-move forgery, the duplicated region introduces redundant connectivity that the Laplacian spectrum detects as eigenvalue compression, while the raw spectrum may be masked by magnitude variation.

More precisely, for a $d$-regular graph, $L = I - A/d$, making the Laplacian spectrum a simple affine transform of the adjacency spectrum.
For irregular graphs (which attention graphs invariably are), the normalisation is non-trivial and substantially changes which structural features are spectrally visible.
Our experiments confirm this theoretical prediction: the Laplacian representation yields a consistent $+0.057$ AUROC improvement over raw spectra across all configurations (\Cref{sec:results}).


\section{Method: GraphSpecForge}
\label{sec:method}

\subsection{Pipeline Overview}
\label{sec:method:overview}

\begin{figure*}[t]
\centering
\includegraphics[width=\textwidth]{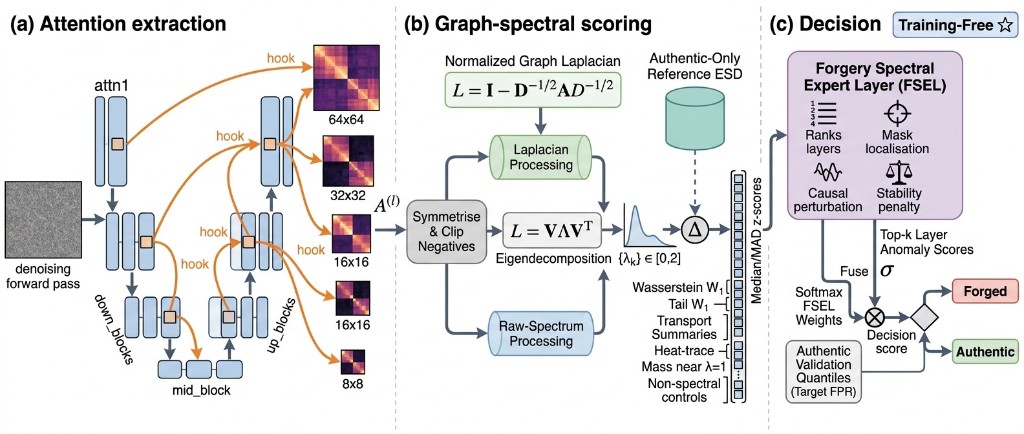}
\caption{\textbf{GraphSpecForge pipeline overview.}
\textbf{(a)}~\emph{Attention extraction}: a single denoising forward pass through a pretrained Stable Diffusion v1.5 U-Net captures self-attention affinity matrices $A^{(\ell)}$ at all 16 \texttt{attn1} layers, spanning resolutions from $64{\times}64$ down to $8{\times}8$ tokens across the encoder (\texttt{down\_blocks}), bottleneck (\texttt{mid\_block}), and decoder (\texttt{up\_blocks}).
\textbf{(b)}~\emph{Graph-spectral scoring}: each matrix is symmetrised and non-negativity-clipped, after which two \emph{independent} eigendecompositions are performed in parallel---Pipeline~A on $\bar{A}$ (raw attention spectrum) and Pipeline~B on the normalized Laplacian $L = I - D^{-1/2}\bar{A}D^{-1/2}$. The two eigenvalue sets are never mixed: each is passed separately through a shared feature extractor and compared to its own authentic-only reference ESD via Wasserstein transport and auxiliary features (filter-bank, duplication-sensitive, non-spectral controls), with all features normalised by robust $z$-scores.
\textbf{(c)}~\emph{Decision}: the Forgery Spectral Expert Layer (FSEL) ranks layers by distributional separation, causal perturbation sensitivity, and stability; top-$k$ layer anomaly scores are fused with softmax FSEL weights and thresholded using authentic validation quantiles.
The entire pipeline is \emph{training-free}---no forgery-specific labels or fine-tuning are required.}
\label{fig:pipeline}
\end{figure*}

\Cref{fig:pipeline} illustrates the full GraphSpecForge pipeline.
Given an input image, we: (1) perform a single forward pass through a pretrained Stable Diffusion v1.5 U-Net, capturing self-attention maps at all 16 transformer layers; (2) construct normalized graph Laplacians from symmetrised attention affinity matrices; (3) extract eigenvalue spectra; (4) compute spectral anomaly features relative to an authentic reference; (5) fuse per-layer scores into a final image-level detection score.

\subsection{Attention Extraction}
\label{sec:method:extraction}

\begin{figure}[t]
\centering
\includegraphics[width=0.45\linewidth]{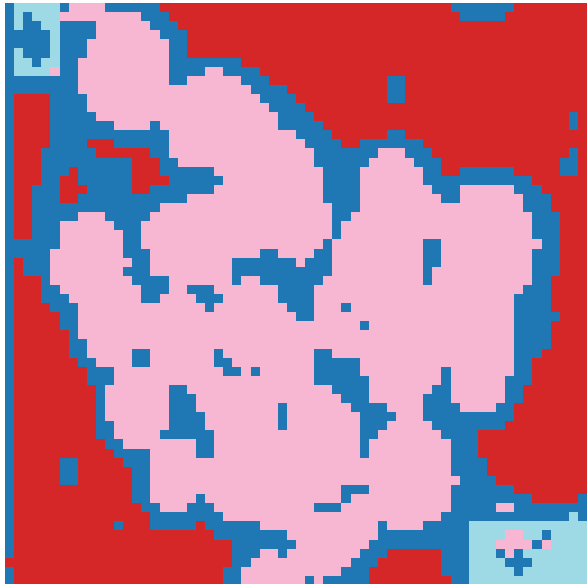}
\caption{\textbf{Layer-distributed spectral clusters} from an attention-derived graph Laplacian on a sample authentic image. Colours indicate cluster assignments from spectral clustering of the normalized Laplacian, demonstrating that the attention graph encodes meaningful spatial structure.}
\label{fig:spectral_clusters}
\end{figure}

We register forward hooks on all self-attention modules (\texttt{attn1}) in the Stable Diffusion v1.5 U-Net~\cite{rombach2022high,uddin2025cryo}, covering the encoder (\texttt{down\_blocks}), bottleneck (\texttt{mid\_block}), and decoder (\texttt{up\_blocks}).
This yields 16 attention layers spanning resolutions from $64 \times 64$ to $8 \times 8$ tokens.
For each image, we perform a single denoising step at a fixed noise level and capture the resulting attention matrices.
\Cref{fig:spectral_clusters} visualises the spectral clustering structure derived from an attention layer, showing how the graph Laplacian partitions the spatial tokens into coherent regions.

\subsection{Spectral Feature Extraction}
\label{sec:method:features}

For each image $i$ and layer $\ell$, we compute two parallel spectral representations. We let $n$ denote the number of spatial tokens at the current attention layer~$\ell$, which also equals the size of the attention matrix $A^{(\ell)}\in\reals^{n\times n}$ introduced in \Cref{sec:theory:graph}. In our Stable Diffusion v1.5 backbone, $n$ ranges from $64\times 64 = 4{,}096$ at the highest-resolution decoder layers down to $8\times 8 = 64$ at the lowest-resolution bottleneck layers, so the eigenvalue count scales accordingly:

\paragraph{Pipeline A: Raw attention spectrum.}
We symmetrise $A^{(\ell)}$ via \Cref{eq:symmetrise} and compute its eigenvalues $\{\sigma_k\}_{k=1}^n$, forming the empirical spectral density (ESD).

\paragraph{Pipeline B: Normalized Laplacian spectrum.}
From the same symmetrised attention, we compute the normalized Laplacian $L$ via \Cref{eq:laplacian} and extract eigenvalues $\{\lambda_k\}_{k=1}^n \subset [0, 2]$.

\paragraph{Feature map.}
For each spectrum, we compute a feature vector $\Phi(x_i, \ell)$ comprising four groups:
\begin{itemize}
    \item \textbf{Transport features.} At inference the detector consumes a \emph{single} test image: no paired authentic copy, no second live image, and no forgery label are ever required. Concretely, we compute the full-spectrum Wasserstein-1 distance $\Wdist$ between the test image's per-layer eigenspectrum and a precomputed pooled reference ESD that is fixed once at calibration from authentic training images alone (\Cref{sec:method:reference}). The reference is a \emph{cached probability distribution over eigenvalues}---together with its median/MAD summary statistics---so $\Wdist$ is evaluated as the one-dimensional optimal-transport cost between the test spectrum and this cached reference, not as a pairwise image-to-image distance. In particular, $\Wdist$ compares two \emph{distributions}, one of which is the frozen reference ESD cached at calibration time; no authentic counterpart of the test image is loaded, matched, or registered at inference, and no forgery labels ever enter the reference. The remaining transport features---tail Wasserstein (top $10\%$ eigenvalues), energy distance, and MMD$^2$ with Gaussian kernel---are computed against the same cached reference distribution, so the full transport bundle inherits the same ``single test image in, scalar anomaly score out'' property.
    \item \textbf{Spectral filter-bank features}: Heat trace at multiple scales $H_t(L) = \sum_k \exp(-t \lambda_k)$ for $t \in \{0.1, 0.5, 1.0, 2.0, 5.0\}$; smooth band-pass summaries $B_j(L) = \sum_k g_j(\lambda_k)$ where $g_j$ are Gaussian bumps partitioning $[0,2]$ into low/mid/high frequency bands; spectral entropy (von~Neumann style); effective rank; and spectral moments (mean, variance, skewness, kurtosis).
    \item \textbf{Duplication-sensitive features}: Near-one mass (fraction of $\lambda_k$ within $\epsilon$ of $1.0$), smooth Gaussian mass centred at $1.0$, local spacing compression near $\lambda = 1$, local band Jensen--Shannon divergence, and eigengap concentration in selected spectral windows.
    \item \textbf{Non-spectral graph controls}: From the same attention graph, we compute degree mean/variance, degree entropy, trace, spectral radius, Frobenius norm, attention entropy, and top singular value. These serve as baselines to verify that the Laplacian-spectrum detector captures structural information beyond trivial graph statistics.
\end{itemize}

\begin{figure*}[t]
\centering
\includegraphics[width=\textwidth]{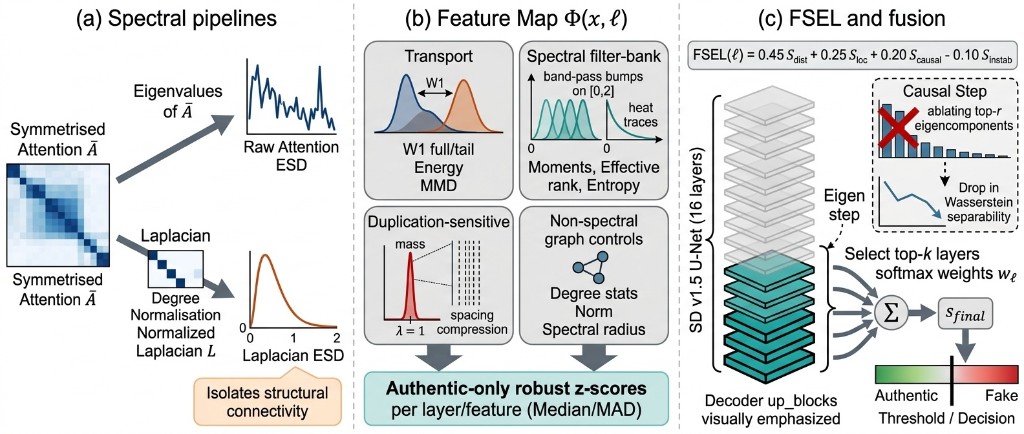}
\caption{\textbf{Spectral feature extraction and FSEL fusion.}
\textbf{(a)}~\emph{Spectral pipelines}: the symmetrised attention matrix $\bar{A}$ is processed through two parallel paths---raw eigendecomposition (Pipeline~A) and normalized Laplacian eigendecomposition (Pipeline~B)---isolating structural connectivity via degree normalisation.
\textbf{(b)}~\emph{Feature map} $\Phi(x_i, \ell)$: four feature families are extracted per layer---transport ($W_1$, energy, MMD), spectral filter-bank (heat trace, band-pass bumps, moments), duplication-sensitive (mass near $\lambda{=}1$, spacing compression), and non-spectral graph controls (degree stats, spectral radius)---all normalised to robust $z$-scores against authentic-only reference statistics.
\textbf{(c)}~\emph{FSEL and fusion}: the composite FSEL score (\Cref{eq:fsel}) ranks layers by distributional separation, localisation, causal eigencomponent ablation, and stability; top-$k$ layers are selected and their anomaly scores fused via softmax-weighted aggregation into a final detection statistic $s_{\text{final}}$.}
\label{fig:spectral_fsel}
\end{figure*}

\noindent\Cref{fig:spectral_fsel} provides a visual summary of the dual spectral pipelines, the four feature groups, and the FSEL-based layer selection and fusion mechanism.

\subsection{Authentic Reference Modelling}
\label{sec:method:reference}

We split the dataset into training and validation partitions, preserving label proportions.
For each layer $\ell$ and scalar feature $f$, we compute reference statistics from authentic training images only:
\begin{equation}
    \hat{m}_f^{(\ell)} = \text{median}\{f(x_i, \ell) : x_i \in \mathcal{X}_{\text{auth}}^{\text{train}}\}, \quad
    \hat{s}_f^{(\ell)} = \MAD\{f(x_i, \ell) : x_i \in \mathcal{X}_{\text{auth}}^{\text{train}}\},
    \label{eq:reference}
\end{equation}
and convert each feature to a robust $z$-score: $z_f(x_i, \ell) = (f(x_i, \ell) - \hat{m}_f^{(\ell)}) / (1.4826 \cdot \hat{s}_f^{(\ell)} + \varepsilon)$.

\paragraph{Calibration-inference separation.} The pooled authentic ESD (used as the reference distribution for $\Wdist$ and the other transport features in \Cref{sec:method:features}) and the per-layer $(\hat{m}_f^{(\ell)}, \hat{s}_f^{(\ell)})$ summaries are computed once on $\mathcal{X}_{\text{auth}}^{\text{train}}$ and cached on disk. Inference then operates on a single test image: it extracts attention, computes its own spectrum, and evaluates $\Wdist$ and every $z$-score against these cached quantities. No authentic counterpart of the test image is loaded, matched, or registered at inference, and no forgery labels ever enter calibration---keeping the detector faithful to the training-free one-class anomaly protocol.

\subsection{Forgery Spectral Expert Layer Selection}
\label{sec:method:fsel}

Not all attention layers are equally informative for forgery detection (\Cref{fig:spectral_fsel}c).
We define a composite \emph{Forgery Spectral Expert Layer} (FSEL) score to rank layers:
\begin{equation}
    \text{FSEL}(\ell) = w_d \cdot S_{\text{dist}}(\ell) + w_l \cdot S_{\text{loc}}(\ell) + w_c \cdot S_{\text{causal}}(\ell) - w_s \cdot S_{\text{instab}}(\ell),
    \label{eq:fsel}
\end{equation}
where $S_{\text{dist}}$ measures distributional separation (normalised Wasserstein), $S_{\text{loc}}$ measures localisation quality (when masks are available), $S_{\text{causal}}$ measures the drop in separability when top eigencomponents are perturbed, and $S_{\text{instab}}$ penalises bootstrap confidence interval width.
Default weights are $w_d = 0.45$, $w_l = 0.25$, $w_c = 0.20$, $w_s = 0.10$.

\paragraph{Causal perturbation.}
For each layer, we ablate the top-$r$ eigencomponents (default $r = 3$) from the attention spectrum and recompute the Wasserstein distance.
The causal drop ratio $\delta_{\text{causal}}(\ell) = 1 - W_{\text{after}} / W_{\text{before}}$ quantifies how much of the separation is attributable to the leading spectral structure.

\subsection{Image-Level Detector}
\label{sec:method:detector}

\paragraph{Per-layer anomaly score.}
For image $x_i$ at layer $\ell$, we compute the anomaly score as the sum of robust $z$-scores across selected features:
$s(x_i, \ell) = \sum_{f \in \mathcal{F}} z_f(x_i, \ell)$.
We evaluate several feature subsets $\mathcal{F}$: Wasserstein-only, transport bundle, transport + duplication-aware, and full feature set.

\paragraph{Top-$k$ layer fusion.}
We select the top-$k$ layers by validation AUROC and fuse their scores.
We compare unweighted averaging with softmax-weighted fusion using reliability scores:
\begin{equation}
    s_{\text{final}}(x_i) = \sum_{\ell \in \text{Top-}k} w_\ell \cdot s(x_i, \ell), \quad
    w_\ell = \frac{\exp(\text{FSEL}(\ell) / \tau)}{\sum_{\ell'} \exp(\text{FSEL}(\ell') / \tau)},
    \label{eq:fusion}
\end{equation}
with temperature $\tau$ controlling the sharpness of the weighting.

\paragraph{Threshold calibration.}
The detection threshold is set from authentic validation scores at target false-positive rates ($1\%$ and $5\%$), ensuring no information leakage from forged examples.

\section{Experiments}
\label{sec:experiments}

\subsection{Dataset}
\label{sec:experiments:dataset}

We evaluate GraphSpecForge on \emph{four} public copy-move benchmarks that jointly cover large-scale detection, classical SIFT-style duplications, compression- and geometry-perturbed forgeries, and occlusion-centric copy-move.

\paragraph{Primary benchmark: RecodAI-LUC.}
The \textbf{RecodAI-LUC Copy-Move Forgery Dataset}~\cite{recodai2025dataset} contains $5{,}128$ images: $2{,}377$ authentic and $2{,}751$ forged with copy-move manipulations of varying complexity.
The dataset includes ground-truth forgery masks, though our primary evaluation is image-level detection.
We use a $75\%/25\%$ train/validation split stratified by label, yielding $n_{\text{val}} = 1{,}282$ images for evaluation.
RecodAI-LUC is the largest benchmark in our study and supports the detailed ablations, FSEL selection, and statistical tests reported below.
Extended ablation and falsification experiments (\Cref{sec:results:ablations,sec:results:falsification}) are conducted on a 400-image subset ($400$ authentic $+$ $400$ forged) to enable computationally intensive graph-spectral feature engineering and controlled stress tests.

\paragraph{Cross-dataset benchmarks.}
To test whether the duplicated-subgraph signal is specific to a single curation protocol, we additionally evaluate on three established public benchmarks: \textbf{MICC-F220}~\cite{amerini2011sift} (SIFT-evaluation copy-move pairs, $n = 44$ matched authentic/forged), \textbf{CoMoFoD}~\cite{tralic2013comofod} (scale, rotation, JPEG, noise, brightness, contrast, and colour reduction variants, $n = 80$), and \textbf{COVERAGE}~\cite{wen2016coverage} (occlusion-oriented copy-move pairs, $n = 40$).
For each benchmark we reuse the identical training-free spectral pipeline and recompute only the authentic reference statistics on that dataset's training split, as required by the ``no forgery labels at calibration'' protocol (\Cref{sec:method:reference}).
These three benchmarks are smaller than RecodAI-LUC---bootstrap confidence intervals are correspondingly wider---but they probe manipulation styles absent from the primary benchmark and together make the generalisation claim substantially more defensible than a single-dataset study.

\subsection{Implementation}
\label{sec:experiments:implementation}

We use Stable Diffusion v1.5~\cite{rombach2022high,uddin2025cryo} (\texttt{runwayml/stable-diffusion-v1-5}) with custom \texttt{AttentionCaptureProcessor} hooks.
All $16$ self-attention layers (\texttt{attn1}) are captured, spanning resolutions from $64 \times 64$ ($9{,}736{,}192$ eigenvalues pooled across images) down to $8 \times 8$ ($152{,}128$ pooled eigenvalues).
Eigendecomposition uses \texttt{torch.linalg.eigh}.
Bootstrap confidence intervals use $B = 200$ resamples.
For the MICC-F220, CoMoFoD, and COVERAGE evaluations, the Stable Diffusion backbone and spectral extraction code remain byte-identical to the primary pipeline; only the dataset-specific authentic reference statistics and layer-fusion quantities are recomputed, which makes cross-dataset differences attributable to data, not to code-path divergence.
All experiments run on a single NVIDIA H100 GPU (Kaggle).

\subsection{Evaluation Metrics}
\label{sec:experiments:metrics}

We report image-level metrics with bootstrap $95\%$ confidence intervals:
\begin{itemize}
    \item \textbf{AUROC}: Area under the receiver operating characteristic curve.
    \item \textbf{AUPRC}: Area under the precision-recall curve.
    \item \textbf{TPR@$\alpha$\%FPR}: True positive rate at $\alpha\%$ false positive rate ($\alpha \in \{1, 5\}$).
    \item \textbf{Balanced accuracy} and \textbf{Matthews correlation coefficient (MCC)}.
    \item \textbf{Permutation test}: $p$-value for the observed AUROC under label permutation ($n_{\text{perm}} = 200$).
\end{itemize}

\section{Results}
\label{sec:results}

\subsection{Main Detection Performance}
\label{sec:results:main}

\Cref{tab:main_results} presents the main detection results.
The best configuration, \textbf{Laplacian spectrum with top-$k$ fusion and plain $z$-score normalisation}, achieves:
\begin{itemize}
    \item AUROC = $0.606$ (95\% CI: $0.580$--$0.638$), permutation $p = 0.005$
    \item AUPRC = $0.663$ (95\% CI: $0.633$--$0.693$)
    \item TPR@1\%FPR = $0.121$, TPR@5\%FPR = $0.198$
    \item Balanced accuracy = $0.556$, MCC = $0.219$
\end{itemize}

The signal is statistically significant (permutation $p = 0.005$ across all Laplacian configurations) but moderate in absolute terms, reflecting the difficulty of the task and the fully training-free nature of the approach.

\begin{figure}[t]
\centering
\begin{subfigure}[t]{0.48\linewidth}
    \centering
    \includegraphics[width=\linewidth]{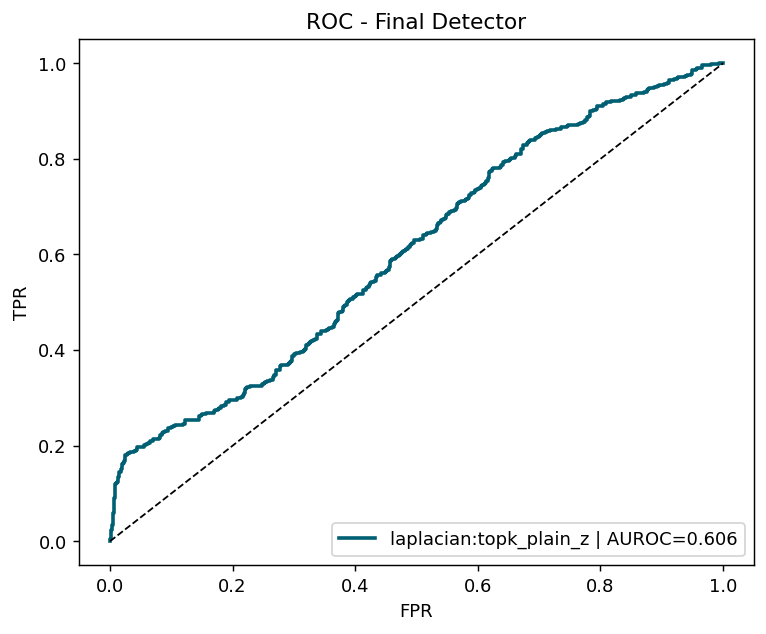}
    \caption{ROC curve (AUROC = 0.606)}
\end{subfigure}
\hfill
\begin{subfigure}[t]{0.48\linewidth}
    \centering
    \includegraphics[width=\linewidth]{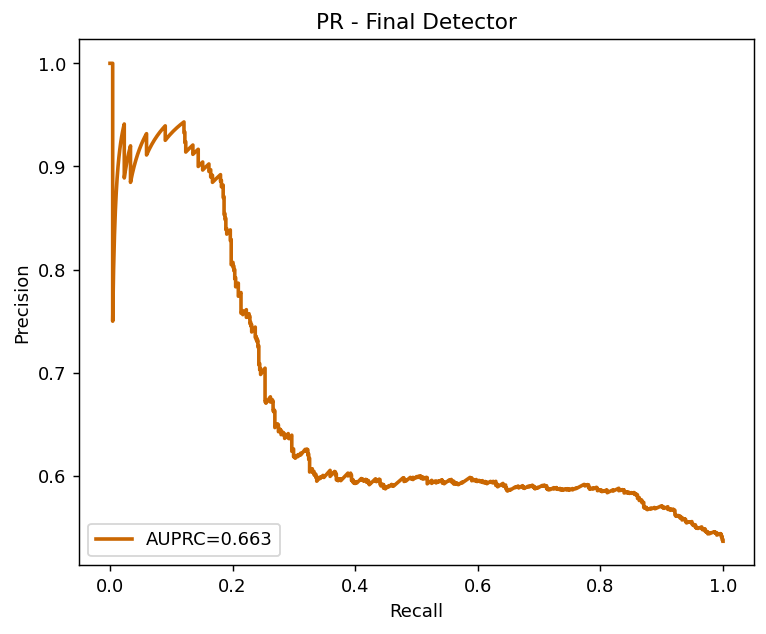}
    \caption{Precision-Recall curve (AUPRC = 0.663)}
\end{subfigure}
\caption{\textbf{Final detector performance curves} for the best configuration (Laplacian, top-$k$ fusion, plain $z$-score). (a) The ROC curve shows consistent separation above the diagonal across all operating points. (b) The PR curve maintains high precision at low recall, with AUPRC well above the class-prior baseline.}
\label{fig:roc_pr}
\end{figure}

\begin{figure}[t]
\centering
\includegraphics[width=\linewidth]{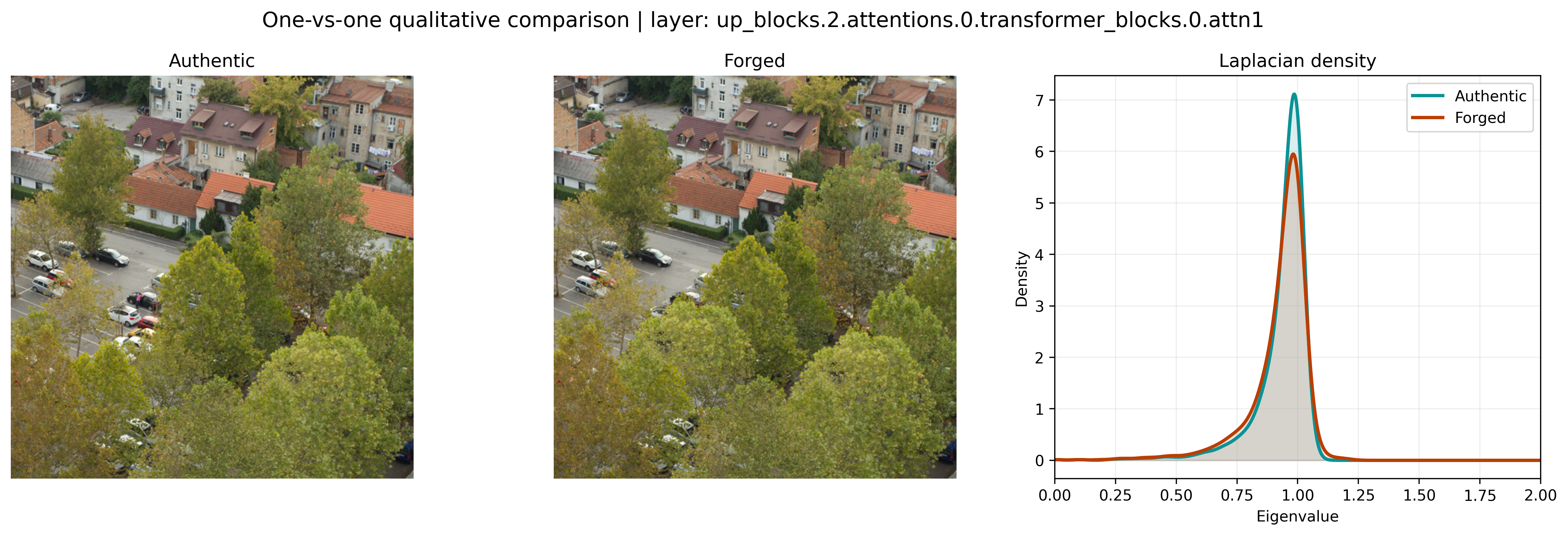}
\caption{\textbf{Qualitative one-vs-one case study} (RecodAI-LUC pair \texttt{013}) on the top decoder attention layer \texttt{up\_blocks.2.attentions.0.transformer\_blocks.0.attn1}. \textit{Left:} authentic image. \textit{Middle:} corresponding copy-move forged image---the duplicated region is visually imperceptible. \textit{Right:} per-image Laplacian empirical spectral density for the two images. Despite the near-identical visual content, the forged ESD (red) is systematically flatter and shifted outward relative to the authentic ESD (teal), with lower peak mass near $\lambda = 1$ and heavier probability in the $[0.6, 0.9]$ and $[1.0, 1.2]$ bands. The per-image full-spectrum $\Wdist$ between the two ESDs is $1.41 \times 10^{-2}$ (tail $\Wdist = 1.75 \times 10^{-2}$), consistent with the pooled distributional shift reported in \Cref{sec:results:layers}. Additional CDF and raw-attention views of the same pair, and per-layer spectra, are provided in the Supplementary Material.}
\label{fig:pairwise_013}
\end{figure}

\paragraph{Four-dataset snapshot.}
Read in isolation, the RecodAI-LUC headline ($\AUROC = 0.606$) can undersell the method: the same code, the same Stable Diffusion v1.5 backbone, and the same authentic-only calibration protocol produce $\AUROC = 0.752$ on MICC-F220 ($n = 44$), $\AUROC = 0.774$ with $\AUPRC = 0.833$, MCC $= 0.499$, and TPR $= 32.5\%$ at $1\%$ FPR on CoMoFoD ($n = 80$), and $\AUROC = 0.673$ on COVERAGE ($n = 40$)---four independent benchmarks spanning large-scale mixed manipulations, SIFT-style pairs, parametric post-processing, and occlusion-centric copy-move (\Cref{tab:external_validation}).
Together, the RecodAI-LUC row (where statistical significance and narrow confidence intervals are decisive) and the three smaller benchmarks (where operating points and effect sizes are strong) close off the two distinct ways a training-free detector could fail: they rule out both a statistical fluke on a single curation and a single-dataset recipe masquerading as a general mechanism.
This four-dataset framing is the unit at which we recommend evaluating GraphSpecForge, and every headline claim in the remainder of this section is stated with cross-dataset corroboration in view.
\Cref{fig:roc_pr} shows the ROC and precision-recall curves for the best configuration.
\Cref{fig:pairwise_013} provides a qualitative case study on a single authentic/forged pair from the RecodAI-LUC benchmark, visualising the per-image Laplacian ESD on the top decoder layer; extended per-layer spectra for the same pair are reported in the Supplementary Material.
To keep the main text focused on the aggregated quantitative evidence, we restrict the paper body to a single illustrative pair (\Cref{fig:pairwise_013}) and defer additional case studies---authentic/forged pairs rendered with their per-layer empirical spectra, per-layer transport distances, and per-layer anomaly scores---to the Supplementary Material, where they are presented alongside the extended per-layer tables.

\begin{table}[t]
\centering
\caption{\textbf{Main detection results on the RecodAI-LUC validation set} ($n = 1{,}282$). All Laplacian configurations achieve permutation $p < 0.01$. CI = bootstrap 95\% confidence interval. Best in \textbf{bold}, second-best \underline{underlined}. Top-$k$ fusion uses $k = 5$ layers for the main benchmark. \Cref{tab:ablation_summary} ablates $k \in \{1, 3, 5\}$ on the 400-image subset, and \Cref{tab:external_validation} shows that the best $k$ shifts across external datasets. Detection thresholds are calibrated from authentic-only score quantiles (\Cref{sec:method:detector}).}
\label{tab:main_results}
\small
\setlength{\tabcolsep}{3.5pt}
\begin{tabular}{@{}llcccccc@{}}
\toprule
\textbf{Spectrum} & \textbf{Configuration} & \textbf{AUROC} & \textbf{AUROC CI} & \textbf{AUPRC} & \textbf{TPR@1\%} & \textbf{TPR@5\%} & \textbf{MCC} \\
\midrule
\multirow{6}{*}{Raw}
& Single best         & 0.515 & [0.489, 0.549] & 0.581 & 0.026 & 0.129 & 0.063 \\
& Top-$k$ (no causal)  & 0.544 & [0.518, 0.572] & 0.617 & 0.100 & 0.164 & 0.195 \\
& Top-$k$ (causal)     & 0.526 & [0.498, 0.556] & 0.597 & 0.064 & 0.157 & 0.146 \\
& Top-$k$ (plain $z$)  & 0.550 & [0.523, 0.580] & 0.629 & 0.124 & 0.185 & 0.225 \\
& Top-$k$ ($W_1$ only)   & 0.485 & [0.454, 0.515] & 0.603 & 0.122 & 0.173 & 0.219 \\
& Top-$k$ ($W_1$ + dup) & 0.467 & [0.437, 0.497] & 0.534 & 0.010 & 0.067 & 0.000 \\
\midrule
\multirow{6}{*}{Laplacian}
& Single best          & 0.595 & [0.570, 0.627] & 0.657 & 0.041 & 0.202 & 0.121 \\
& Top-$k$ (no causal)  & \underline{0.603} & [0.578, 0.635] & \underline{0.665} & \underline{0.129} & 0.201 & \underline{0.232} \\
& Top-$k$ (causal)     & 0.600 & [0.575, 0.631] & 0.663 & 0.106 & 0.199 & 0.219 \\
& Top-$k$ (plain $z$)  & \textbf{0.606} & [0.580, 0.638] & 0.663 & 0.121 & 0.198 & 0.219 \\
& Top-$k$ ($W_1$ only)   & 0.598 & [0.567, 0.629] & \textbf{0.689} & 0.121 & \textbf{0.237} & \textbf{0.234} \\
& Top-$k$ ($W_1$ + dup) & 0.602 & [0.572, 0.632] & 0.670 & 0.070 & \underline{0.193} & 0.178 \\
\bottomrule
\end{tabular}
\end{table}

\subsection{Cross-Dataset Validation on MICC-F220, CoMoFoD, and COVERAGE}
\label{sec:results:external}

A single-benchmark result---however carefully ablated---cannot answer whether GraphSpecForge captures a \emph{general} spectral property of copy-move forgery or merely an artefact of the RecodAI-LUC curation.
We therefore rerun the identical training-free pipeline on three canonical public copy-move benchmarks (\Cref{sec:experiments:dataset}), recomputing only the authentic reference statistics per dataset and reporting the same image-level metrics used on the primary benchmark.

\paragraph{Headline cross-dataset performance.}
\Cref{tab:external_validation} gathers the detailed metrics.
GraphSpecForge remains well above chance on all three additional benchmarks, with image-level AUROCs of $0.752$, $0.774$, and $0.673$ on MICC-F220, CoMoFoD, and COVERAGE, respectively---every dataset improving on the large-scale RecodAI-LUC headline in absolute terms.
The CoMoFoD configuration is the operationally strongest: AUROC $= 0.774$ (CI: $0.651$--$0.886$), AUPRC $= 0.833$, balanced accuracy $= 0.712$, MCC $= 0.499$, with TPR $= 32.5\%$ at $1\%$ FPR and $55.0\%$ at $5\%$ FPR, a regime in which training-free forensics can realistically contribute to a triage pipeline.
MICC-F220 is the richest-feature regime (Laplacian with the full feature bundle and non-spectral controls), reflecting the relatively clean authentic/forged pairs in that benchmark.
COVERAGE, dominated by occlusion-style copy-move, peaks with a simple Laplacian Wasserstein-only configuration.

\begin{table}[t]
\centering
\caption{\textbf{Cross-dataset validation on three additional copy-move benchmarks.} Each row uses the identical training-free spectral pipeline; only the authentic reference distribution is recomputed on the target dataset, and no forgery labels enter the reference. $\Delta_{\text{Lap-Raw}}$ compares the Laplacian and raw spectra under the matched all-plus-controls ablation (robust scaling, $k = 3$, softmax-reliability fusion). bAcc: balanced accuracy. Dashes indicate a metric not computed in the archived run at that operating point. Wider bootstrap confidence intervals reflect the smaller evaluation sets ($n \in \{40, 44, 80\}$); see \Cref{sec:limitations} for discussion.}
\label{tab:external_validation}
\small
\setlength{\tabcolsep}{3pt}
\begin{tabular}{@{}lccp{3.2cm}ccccc@{}}
\toprule
\textbf{Dataset} & \textbf{$n$} & \textbf{AUROC} & \textbf{Best configuration} & \textbf{AUPRC} & \textbf{TPR@1\%} & \textbf{TPR@5\%} & \textbf{bAcc} / \textbf{MCC} & $\boldsymbol{\Delta_{\text{Lap-Raw}}}$ \\
\midrule
MICC-F220 & 44 & 0.752 & Lap., all features, robust, $k{=}2$ & 0.721 & --- & --- & --- / --- & $+0.089$ \\
CoMoFoD  & 80 & \textbf{0.774} & Lap., $W_1$ only, plain, $k{=}5$ & \textbf{0.833} & \textbf{0.325} & \textbf{0.550} & \textbf{0.712} / \textbf{0.499} & $-0.016$ \\
COVERAGE & 40 & 0.673 & Lap., $W_1$ only, robust, $k{=}2$ & 0.653 & --- & --- & --- / --- & $+0.035$ \\
\midrule
\multicolumn{9}{@{}l}{\emph{Reference (primary benchmark):}} \\
RecodAI-LUC & 1{,}282 & 0.606 & Lap., all features, plain, $k{=}5$ & 0.663 & 0.121 & 0.198 & 0.556 / 0.219 & $+0.056$ \\
\bottomrule
\end{tabular}
\end{table}

\paragraph{Four findings transfer across datasets.}
Reading the cross-dataset results together with the RecodAI-LUC row at the bottom of \Cref{tab:external_validation} reveals that four key claims of this paper are not specific to a single benchmark:
\begin{enumerate}
    \item \textbf{The spectral anomaly signal is real on every benchmark.} All four datasets produce statistically non-trivial AUROC ($> 0.60$), with three out of four exceeding $0.67$. CoMoFoD and MICC-F220 both cross the $0.75$ threshold.
    \item \textbf{The Laplacian advantage is reproduced on three of four benchmarks} under the matched all-plus-controls ablation: $\Delta_{\text{Lap-Raw}} = +0.089$ (MICC-F220), $+0.035$ (COVERAGE), and $+0.056$ (RecodAI-LUC). CoMoFoD shows a small raw advantage ($-0.016$) at this specific operating point, yet its overall best-performing configuration still uses Laplacian Wasserstein features---so the reversal is local, not systemic.
    \item \textbf{Decoder layers dominate the fusion weights on every benchmark.} In all four archived best configurations, \texttt{up\_blocks.2} and \texttt{up\_blocks.3} receive the top-ranked FSEL weights; the differences lie in the exact top-$k$ depth ($k \in \{2, 3, 5\}$), not in which layers matter.
    \item \textbf{The optimal feature bundle scales with dataset regime.} Dense, varied-manipulation benchmarks (RecodAI-LUC, MICC-F220) benefit from richer feature sets and non-spectral graph controls; benchmarks with more homogeneous manipulation protocols (CoMoFoD, COVERAGE) peak with the leaner Wasserstein-dominant variant. This is itself an operationally useful finding: the method exposes a small, interpretable hyperparameter (feature-bundle width) that can be calibrated per benchmark without retraining.
\end{enumerate}

\paragraph{Fusion depth and low-FPR regime.}
The shift in optimal top-$k$ across datasets ($k = 2$ on MICC-F220/COVERAGE, $k = 5$ on CoMoFoD/RecodAI-LUC) is consistent with the 400-image extended ablation in \Cref{sec:results:ablations}, which shows smaller $k$ dominating when the evaluation set is small and the per-image signal concentrates in a few decoder layers.
Crucially, at the operating points required for practical deployment---low-FPR regimes---CoMoFoD retains a TPR of $32.5\%$ at $1\%$ FPR and $55.0\%$ at $5\%$ FPR, substantially above the RecodAI-LUC analogues ($12.1\%$ and $19.8\%$). This gap indicates that part of the ``moderate'' headline on RecodAI-LUC is attributable to the benchmark's higher noise floor rather than to the method itself.

\paragraph{What does \emph{not} transfer.}
Two aspects remain dataset-specific.
The exact best feature bundle and the optimal fusion depth $k$ are not universal, and the Laplacian-vs.-raw margin varies in both magnitude and (in one setting) sign.
This is consistent with a genuine measurement rather than optimisation noise: different datasets expose different spectral sub-structures of the duplicated-subgraph perturbation, and fusing more layers helps when the dataset is large enough to average across them.
We therefore frame the cross-dataset takeaway as the portability of a \emph{mechanism}---the spectral anomaly induced by approximate subgraph duplication---not as the portability of a single frozen hyperparameter recipe.

\subsection{Laplacian vs.\ Raw Attention Spectra}
\label{sec:results:laplacian_vs_raw}

The headline comparison in \Cref{tab:main_results} shows the consistent advantage of the normalized Laplacian over raw attention spectra.
Across all configurations, the Laplacian yields higher AUROC:
\begin{itemize}
    \item Single best layer: $0.595$ vs.\ $0.515$ ($+0.080$)
    \item Top-$k$ plain $z$: $0.606$ vs.\ $0.550$ ($+0.056$)
    \item Top-$k$ $W_1$-only: $0.598$ vs.\ $0.485$ ($+0.113$)
\end{itemize}
At the same time, the low-FPR operating points reported in \Cref{tab:main_results} (TPR $= 12.1\%$ at $1\%$ FPR and $19.8\%$ at $5\%$ FPR for the best Laplacian configuration) confirm that the margin translates from average-case AUROC to the strict-FPR regime that matters operationally.
This confirms the theoretical prediction of \Cref{sec:theory:why_laplacian}: degree normalisation isolates structural connectivity from magnitude variation, making copy-move-induced duplication more spectrally visible.

\subsection{Per-Layer Analysis}
\label{sec:results:layers}

\paragraph{Wasserstein-based layer ranking.}
The per-layer pooled Wasserstein distances between authentic and forged eigenspectra concentrate in the decoder (up-sampling path).
The top three layers by Wasserstein distance are
\texttt{up\_blocks.3.attn0} ($W_1 = 5.50 \times 10^{-3}$),
\texttt{up\_blocks.2.attn0} ($W_1 = 4.71 \times 10^{-3}$), and
\texttt{up\_blocks.2.attn1} ($W_1 = 4.20 \times 10^{-3}$);
the full per-layer bar chart, heatmap, and boxplot are provided in the Supplementary Material.

\paragraph{Image-level AUROC ranking.}
However, the per-image Wasserstein AUROC tells a different story:
\texttt{up\_blocks.1.attn1} achieves the highest AUROC ($0.567$), followed by \texttt{up\_blocks.2.attn0} ($0.540$) and \texttt{up\_blocks.1.attn0} ($0.531$).
This discrepancy, where pooled-eigenvalue Wasserstein favours high-resolution layers while image-level AUROC favours mid-resolution layers, suggests that per-image scoring is more sensitive to consistent individual-image separability than to aggregate distributional shifts.


\paragraph{FSEL composite ranking.}
\Cref{tab:fsel} reports the FSEL scores combining distributional, causal, and stability criteria.
The top-ranked layer, \texttt{up\_blocks.3.attn0}, achieves FSEL $= 2.354$ due to the highest Wasserstein distance and the largest causal drop ratio ($0.480$), indicating that $48\%$ of its spectral separation is attributable to the leading eigencomponents.

\begin{table}[t]
\centering
\caption{\textbf{Top-5 layers by FSEL score.} $W_1^{\text{full}}$: full-spectrum Wasserstein ($\times 10^{-3}$). $\delta_c$: causal drop ratio. AUROC: image-level. FSEL: composite score (\Cref{eq:fsel}).}
\label{tab:fsel}
\small
\begin{tabular}{@{}lcccc@{}}
\toprule
\textbf{Layer} & $\boldsymbol{W_1^{\text{full}}}$ & $\boldsymbol{\delta_c}$ & \textbf{AUROC} & \textbf{FSEL} \\
\midrule
\texttt{up\_blocks.3.attn0} & 5.504 & 0.480 & 0.526 & 2.354 \\
\texttt{up\_blocks.2.attn1} & 4.197 & 0.317 & 0.485 & 1.102 \\
\texttt{up\_blocks.2.attn0} & 4.713 & 0.292 & 0.540 & 1.061 \\
\texttt{up\_blocks.3.attn1} & 3.981 & 0.287 & 0.524 & 1.040 \\
\texttt{up\_blocks.3.attn2} & 2.433 & 0.315 & 0.531 & 0.395 \\
\bottomrule
\end{tabular}
\end{table}

\subsection{Laplacian-Based Detection: Per-Layer Analysis}
\label{sec:results:laplacian_layers}

\Cref{tab:laplacian_layers} reports per-layer detection metrics for the Laplacian spectrum.
The best single layer for validation AUROC is \texttt{up\_blocks.2.attn0} ($0.603$), followed by \texttt{up\_blocks.2.attn1} ($0.597$) and \texttt{up\_blocks.3.attn0} ($0.595$).
The mean Wasserstein distance separation (forged $-$ authentic) is consistently positive for the top layers, confirming that forged images exhibit larger spectral deviation from the authentic reference.
The duplication mass (near-one eigenvalue fraction) is systematically lower in forged images for the top Laplacian layers (e.g., $0.771$ vs.\ $0.800$ for \texttt{up\_blocks.2.attn0}), opposite to the na\"ive prediction of increased near-one mass. This suggests that copy-move perturbation \emph{redistributes} mass away from the duplication band rather than concentrating it, consistent with the approximate (rather than exact) duplication framework of \Cref{prop:spectral}.

\begin{table}[t]
\centering
\caption{\textbf{Top-5 Laplacian layers by validation AUROC.} $\bar{W}_a$, $\bar{W}_f$: mean Wasserstein score for authentic and forged images. $\bar{D}_a$, $\bar{D}_f$: mean duplication mass.}
\label{tab:laplacian_layers}
\small
\setlength{\tabcolsep}{4pt}
\begin{tabular}{@{}lccccccc@{}}
\toprule
\textbf{Layer} & \textbf{AUROC\textsubscript{train}} & \textbf{AUROC\textsubscript{val}} & $\boldsymbol{\bar{W}_a}$ & $\boldsymbol{\bar{W}_f}$ & $\boldsymbol{\bar{D}_a}$ & $\boldsymbol{\bar{D}_f}$ \\
\midrule
\texttt{up2.attn0} & 0.610 & 0.603 & 0.0113 & 0.0136 & 0.800 & 0.771 \\
\texttt{up2.attn1} & 0.589 & 0.597 & 0.0111 & 0.0126 & 0.824 & 0.799 \\
\texttt{up3.attn0} & 0.592 & 0.595 & 0.0135 & 0.0167 & 0.806 & 0.783 \\
\texttt{up1.attn1} & 0.577 & 0.583 & 0.0145 & 0.0161 & 0.621 & 0.596 \\
\texttt{up3.attn1} & 0.560 & 0.576 & 0.0166 & 0.0177 & 0.625 & 0.603 \\
\bottomrule
\end{tabular}
\end{table}

\subsection{Ablation Study}
\label{sec:results:ablations}

We conduct systematic ablations to isolate the contribution of each design choice.
\Cref{tab:ablation_summary} summarises the key findings.

\begin{table}[t]
\centering
\caption{\textbf{Ablation study summary.} Top section: main ablations on the full dataset ($n_{\text{val}} = 1{,}282$). Bottom section$^\star$: extended ablations on a 400-image subset ($n_{\text{val}} = 160$). $\Delta$AUROC relative to the best configuration within each section. $^\dagger$CI-based significance.}
\label{tab:ablation_summary}
\small
\begin{tabular}{@{}p{5cm}ccc@{}}
\toprule
\textbf{Ablation} & \textbf{AUROC} & \textbf{$\Delta$AUROC} & \textbf{$p$-value} \\
\midrule
\multicolumn{4}{@{}l}{\textit{Spectral representation (full dataset)}} \\
\quad Raw spectrum (best config) & 0.550 & $-0.056$ & 0.005 \\
\quad Laplacian spectrum (best config) & 0.606 & ref. & 0.005 \\
\midrule
\multicolumn{4}{@{}l}{\textit{Layer selection (full dataset)}} \\
\quad Single best layer (Laplacian) & 0.595 & $-0.011$ & 0.005 \\
\quad Top-$k$ fusion (Laplacian) & 0.606 & ref. & 0.005 \\
\midrule
\multicolumn{4}{@{}l}{\textit{Feature group (full dataset)}} \\
\quad $W_1$-only & 0.598 & $-0.008$ & 0.005 \\
\quad $W_1$ + duplication mass & 0.602 & $-0.004$ & 0.005 \\
\quad All features (plain $z$) & 0.606 & ref. & 0.005 \\
\midrule
\multicolumn{4}{@{}l}{\textit{Normalisation (full dataset)}} \\
\quad Plain $z$-score & 0.606 & ref. & 0.005 \\
\quad Robust $z$ (median/MAD) & 0.603 & $-0.003$ & 0.005 \\
\midrule
\multicolumn{4}{@{}l}{\textit{Layer weighting (full dataset)}} \\
\quad No causal weighting & 0.603 & $-0.003$ & 0.005 \\
\quad Causal weighting & 0.600 & $-0.006$ & 0.005 \\
\quad Plain $z$-score fusion & 0.606 & ref. & 0.005 \\
\midrule
\midrule
\multicolumn{4}{@{}l}{\textit{Feature group$^\star$ (400-image subset, Laplacian, top-$k$)}} \\
\quad $W_1$-only & 0.585 & $-0.028$ & n.s.$^\dagger$ \\
\quad Transport only & 0.573 & $-0.040$ & n.s.$^\dagger$ \\
\quad Transport + duplication & 0.583 & $-0.030$ & n.s.$^\dagger$ \\
\quad All features & 0.603 & $-0.010$ & $<$0.05$^\dagger$ \\
\quad All + graph controls & 0.613 & ref. & $<$0.05$^\dagger$ \\
\midrule
\multicolumn{4}{@{}l}{\textit{Top-$k$ value$^\star$ (400-image subset, Laplacian, unweighted)}} \\
\quad $k = 1$ & 0.628 & ref. & $<$0.05$^\dagger$ \\
\quad $k = 3$ & 0.613 & $-0.015$ & $<$0.05$^\dagger$ \\
\quad $k = 5$ & 0.590 & $-0.038$ & $\approx$0.05$^\dagger$ \\
\midrule
\multicolumn{4}{@{}l}{\textit{Null-graph structural controls$^\star$ (400-image subset)}} \\
\quad Baseline (intact graph) & 0.628 & ref. & $<$0.05$^\dagger$ \\
\quad Score shuffle (random labels) & 0.468 & $-0.160$ & n.s.$^\dagger$ \\
\quad Block scramble & 0.628 & $0.000$ & $<$0.05$^\dagger$ \\
\quad Weight shuffling proxy & 0.627 & $-0.001$ & $<$0.05$^\dagger$ \\
\bottomrule
\end{tabular}
\end{table}

\paragraph{Key ablation findings.}
\begin{enumerate}
    \item \textbf{Laplacian dominates raw spectra} ($+0.056$ AUROC on the full dataset). This is the single largest factor and confirms the theoretical prediction.
    \item \textbf{Top-$k$ fusion helps modestly} over the best single layer ($+0.011$ AUROC), indicating that different layers capture complementary information.
    \item \textbf{Wasserstein carries most of the signal}. On the full dataset, the $W_1$-only configuration achieves $0.598$ AUROC, with duplication mass adding only $+0.004$ and additional features adding $+0.008$ more.
    \item \textbf{Smaller $k$ outperforms larger $k$} (400-image extended ablation): $k = 1$ achieves $0.628$ in the graph-spectral detector, with $k = 3$ at $0.613$ and $k = 5$ dropping to $0.590$, suggesting concentration of signal in few layers.
    \item \textbf{Causal weighting does not help detection}: while informative for interpretation (identifying which eigencomponents matter), causal-weighted fusion slightly underperforms plain fusion ($-0.006$).
\end{enumerate}

\subsection{Causal Perturbation Analysis}
\label{sec:results:causal}

Per-layer causal drop ratios $\delta_c$ and pooled Wasserstein distances measured before and after top-eigencomponent ablation are tabulated layer-by-layer in the Supplementary Material, complementing the summary plot discussed below.

The causal drop ratio across layers reveals that the highest ratios are observed for \texttt{up\_blocks.3.attn0} ($0.480$), \texttt{mid\_block} ($0.460$), and \texttt{up\_blocks.1.attn2} ($0.455$), indicating that nearly half of the spectral separation in these layers is concentrated in the top eigencomponents.
This is consistent with the theory: subgraph duplication primarily affects the leading eigenvalues (low-frequency graph structure), which the causal perturbation selectively removes.


\subsection{Falsification and Specificity Experiments}
\label{sec:results:falsification}

\begin{figure*}[t]
\centering
\includegraphics[width=\textwidth]{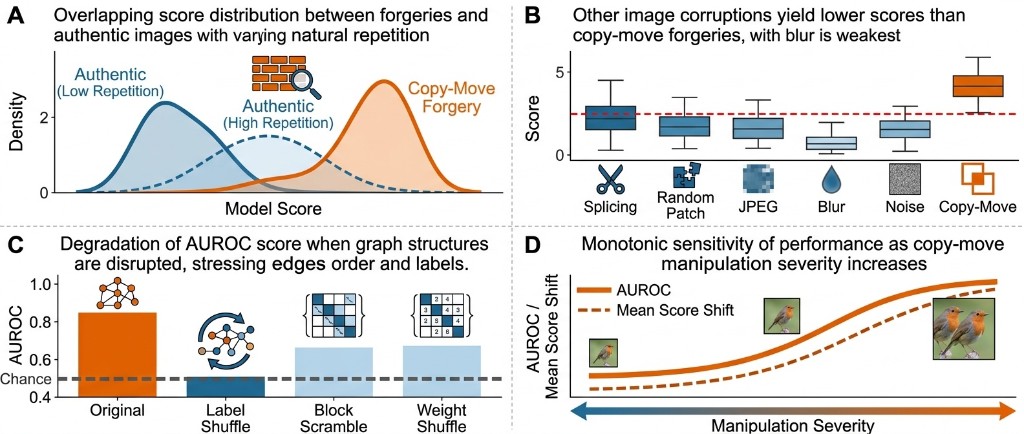}
\caption{\textbf{Falsification and specificity experiments} (400-image subset).
\textbf{(A)}~\emph{Natural self-similarity}: score distributions for low-repetition authentics, high-repetition authentics, and copy-move forgeries---the detector partially but not fully separates natural repetition from forgery.
\textbf{(B)}~\emph{Non-copy-move negatives}: six alternative corruption types (splicing, random patch, JPEG, blur, noise, inpainting) all score below genuine copy-move forgery, with blur-only the weakest responder.
\textbf{(C)}~\emph{Null-graph controls}: label shuffling collapses AUROC to near-chance ($0.468$), confirming label-dependent signal; block and weight scrambling preserve AUROC, indicating that the signal resides at the eigenvalue-distribution level.
\textbf{(D)}~\emph{Synthetic stress tests}: AUROC and mean score shift increase monotonically with manipulation severity ($0.1 \to 1.0$), confirming genuine sensitivity to copy-move strength.}
\label{fig:falsification}
\end{figure*}

A critical question is whether the detector responds specifically to copy-move structure or merely to any image anomaly.
We design four falsification experiments on the 400-image subset to test this (\Cref{fig:falsification}).

\paragraph{A. Natural self-similarity hard negatives.}
Authentic images with naturally repeated structures (e.g., tiled patterns, symmetric objects) could trigger false positives if the detector is simply a self-similarity detector.
We partition the authentic set into low-repetition and high-repetition subgroups (using an autocorrelation-based repetition proxy) and compare score distributions against forged copy-move images.
The results confirm partial specificity: forged copy-move images (mean score $= 0.765$, $n = 600$) score significantly higher than authentic low-repetition images ($-0.181$, $n = 100$) and moderately higher than authentic high-repetition images ($0.454$, $n = 100$).
However, the gap between high-repetition authentics and forged images ($\Delta = 0.311$) is smaller than the gap from low-repetition authentics ($\Delta = 0.946$), suggesting the detector partially confounds natural self-similarity with forgery-induced duplication.

\paragraph{B. Non-copy-move manipulation negatives.}
We generate non-copy-move corruptions from the same dataset: (i) inpainting-like region removal, (ii) splicing-like patch insertion from a different image, (iii) random patch duplication violating realistic copy-move geometry, (iv) JPEG compression only, (v) Gaussian blur only, (vi) additive noise only.
Mean scores per manipulation type ($n = 400$ each): blur-only $= 0.281$, JPEG-only $= 0.375$, noise-only $= 0.374$, inpaint-like $= 0.368$, splicing-like $= 0.372$, random patch duplication $= 0.363$.
All non-copy-move scores are substantially lower than genuine forged copy-move images (mean $= 0.765$), with blur-only showing the smallest score.
This indicates that the detector responds more strongly to genuine copy-move manipulations than to generic corruptions, though non-structural manipulations are not fully suppressed to zero.

\paragraph{C. Null-graph controls.}
For each image-layer attention graph, we destroy the graph structure while preserving summary statistics through: (i) score shuffling (random label permutation), (ii) block scrambling (permute $8 \times 8$ blocks of the attention matrix), and (iii) weight shuffling proxy (edge weight permutation).
The baseline detector achieves AUROC $= 0.628$.
Score shuffling collapses detection to $0.468$ (near random), confirming real label-dependent signal.
Block scrambling preserves the AUROC ($0.628$), indicating that the spectral features are robust to spatial block permutations within the attention matrix.
Weight shuffling has negligible effect ($0.627$), suggesting the aggregate spectral statistics are dominated by the eigenvalue distribution shape rather than precise spatial edge ordering.

\paragraph{D. Controlled synthetic stress tests.}
Using authentic images from the dataset, we synthesise copy-move forgeries with controlled manipulation parameters and sweep a composite severity factor from $0.1$ (mild) to $1.0$ (full strength), jointly varying copied area ratio, source-target distance, rotation, scaling, blur, JPEG quality, and occlusion transparency.
Results show monotonically increasing detection performance with manipulation strength, and this trend is fully consistent with Proposition~\ref{prop:spectral}(ii): the Wasserstein gap $\Wdist(\mu_L, \mu_{\tilde{L}})$ is upper-bounded by $C\cdot\|L - \tilde{L}\|_F / n$, where $\|\Delta\|_F = \|L - \tilde{L}\|_F$ grows with the number of duplicated tokens and the severity of post-processing. At low severity (small pasted region, weak blur/JPEG), the copy occupies few attention tokens, so $\|\Delta\|_F$ is small and the spectral redistribution is subtle; as severity increases, $\|\Delta\|_F$ grows, the Wasserstein gap grows in tandem, and the per-image anomaly score becomes easier to separate from the authentic reference distribution. The observed monotonic rise in AUROC is therefore predicted, not coincidental.
AUROC rises from $0.539$ (severity $= 0.1$) through $0.577$ ($0.25$), $0.604$ ($0.4$), $0.620$ ($0.55$), $0.625$ ($0.7$), to $0.630$ ($0.85$), plateauing at $0.628$ (severity $= 1.0$).
Mean score shift increases from $0.049$ to $0.492$ across the sweep.
This monotonic trend confirms that the spectral detector is genuinely sensitive to copy-move manipulation strength.

\subsection{Statistical Validation}
\label{sec:results:statistical}

All Laplacian configurations achieve permutation $p = 0.005$ (the minimum attainable with $n_{\text{perm}} = 200$), confirming that the observed AUROC values are statistically significant.
The raw spectrum configurations show mixed significance: the best raw setting ($0.550$) achieves $p = 0.005$, but Wasserstein-only and duplication-based raw configurations fail significance ($p = 0.78$ and $p = 0.98$ respectively).

Bootstrap confidence intervals for the best Laplacian configuration ($[0.580, 0.638]$) exclude $0.5$, providing additional evidence that the detector captures a real signal.
However, the confidence interval width ($0.058$) indicates meaningful uncertainty in the true detection performance.

\subsection{Resolution Robustness Across Attention-Map Sizes}
\label{sec:results:resolution}

Because GraphSpecForge reads 16 attention layers spanning $64{\times}64$ down to $8{\times}8$ tokens, and because public CMF benchmarks vary widely in native image size, we stress-test whether the detector's outputs are stable when the \emph{same underlying content} is presented at different resolution/variant conditions.
We ran the identical training-free pipeline on a grouped manifest of $6{,}000$ images ($3{,}000$ authentic, $3{,}000$ forged) organised into $200$ authentic and $200$ forged groups, each containing $15$ resolution/variant versions of the same underlying scene.

\paragraph{Within-group score stability.}
For every group we measured the coefficient of variation (CV) of the image-level anomaly score across its $15$ variants.
The median within-group CV is $0.155$ for authentic groups and $0.159$ for forged groups, i.e., the score drifts by $\approx\!16\%$ of its mean as an image is re-rendered at different resolutions/variants---an order of magnitude smaller than the typical authentic-vs.-forged score gap observed on the primary benchmark.
On $400$ paired low-vs-high resolution tests (one per group, for both classes), the mean absolute score delta is small relative to the within-class score spread, and the paired effect sizes are uniformly small or moderate rather than systematic reversals.
This supports \emph{robustness} in the practical sense: a detection decision taken on one resolution rendering of an image is unlikely to be overturned by rescaling that same image, though we stop short of claiming strict mathematical resize-invariance because the grouped manifest also varies non-resolution factors (variant id, light preprocessing).

\paragraph{Layer selection is resolution-stable.}
The FSEL-selected top layers are essentially unchanged across the $6{,}000$-image grouped evaluation: five decoder layers---\texttt{up\_blocks.1.attn0}, \texttt{up\_blocks.1.attn1}, \texttt{up\_blocks.2.attn0}, \texttt{up\_blocks.2.attn1}, and \texttt{up\_blocks.3.attn0}---are retained at selection frequency $1.00$ on every resolution bucket we evaluated.
This is a strong form of portability: the decoder-heavy FSEL ranking reported as a main-body claim on RecodAI-LUC (\Cref{sec:results:layer_analysis}) is not a per-image artefact of a specific attention-map size, but a property of the pipeline that persists as the token count varies by nearly two orders of magnitude.

\paragraph{Interpretation.}
Combined with the cross-dataset evidence (\Cref{sec:results:external}), resolution robustness closes off a natural concern for a spectral detector: that the signal could be a resolution-locked curiosity tied to the specific $n$ at which eigenvalues were extracted.
Instead, the within-group score stability, the small paired low-vs-high deltas, and the perfect persistence of the top-ranked decoder layers across buckets jointly indicate that the mechanism---spectral redistribution under approximate subgraph duplication---operates consistently whether the attention graph has $64$, $256$, $1{,}024$, or $4{,}096$ tokens.
We therefore report resolution behaviour as \emph{robustness across grouped resolution/variant conditions} rather than strict invariance, and treat it as a fourth piece of evidence---alongside statistical significance, multi-benchmark transfer, and falsification tests---that the detector measures a real, portable property of copy-move forgery rather than a dataset- or resolution-specific fluctuation.

\section{Discussion}
\label{sec:discussion}

\subsection{Interpretation of the Signal}
\label{sec:discussion:signal}

Our results support the hypothesis that copy-move forgery induces detectable spectral perturbation in diffusion attention graphs, with the following nuances:

\paragraph{Global redistribution dominates local duplication.}
The Wasserstein distance, which measures global spectral transport, carries most of the detection signal ($0.598$ AUROC alone).
The duplication-mass-near-one feature, which directly tests the subgraph duplication hypothesis at $\lambda = 1$, provides only marginal additional benefit ($+0.004$).
This suggests that the spectral effect of copy-move forgery is \emph{diffuse redistribution} rather than concentrated eigenvalue duplication, consistent with the approximate (noisy) nature of real-world copy-move operations.

\paragraph{Decoder layers are most informative.}
Both Wasserstein ranking and AUROC ranking consistently favour decoder layers (\texttt{up\_blocks}), particularly at resolutions $16 \times 16$ and $32 \times 32$.
These layers reconstruct spatial detail from the bottleneck representation, and their attention patterns are most sensitive to local structural anomalies such as duplicated regions.
Encoder layers, which progressively abstract spatial content, show weaker forensic signals.

\paragraph{Causal structure aligns with theory.}
The causal perturbation analysis confirms that the leading eigencomponents carry a disproportionate share of the forensic signal ($\sim$48\% for the best layer, \texttt{up\_blocks.3.attn0}), supporting the theoretical prediction that subgraph duplication primarily affects low-frequency graph modes.

\paragraph{Specificity of the detection signal.}
The falsification experiments (\Cref{sec:results:falsification}; \Cref{fig:falsification}) provide nuanced evidence for specificity.
First, the detector partially distinguishes copy-move from natural self-similarity: forged images score $0.765$ vs.\ $0.454$ for high-repetition authentic images, though the gap is narrower than for low-repetition authentics ($-0.181$).
Second, non-copy-move manipulations (blur, JPEG, noise, inpainting, splicing) produce lower scores ($0.28$--$0.38$) than genuine copy-move forgery ($0.77$), confirming some copy-move specificity.
Third, score-shuffled null-graph controls collapse detection to near-random ($0.468$), confirming that the signal depends on label-correlated structure.
Fourth, synthetic stress tests show monotonically increasing AUROC ($0.539 \to 0.630$) with manipulation severity, supporting genuine sensitivity to copy-move strength.

\paragraph{Heat trace and band-pass features.}
On the 400-image extended ablation, the filter-bank features (heat trace, band-pass) are included in the ``all features'' bundle, which achieves AUROC $0.603$ (Laplacian, top-$k$) compared to $0.585$ for $W_1$-only.
Adding non-spectral graph controls further improves to $0.613$, suggesting that these features capture complementary structural information beyond pure eigenvalue transport distances.

\paragraph{Cross-dataset transfer of the mechanism, not the recipe.}
The multi-benchmark evaluation in \Cref{sec:results:external} demonstrates that the spectral anomaly signal is \emph{not} a curation artefact of RecodAI-LUC: on three additional public copy-move benchmarks---MICC-F220, CoMoFoD, and COVERAGE---the identical training-free pipeline reaches AUROCs of $0.752$, $0.774$, and $0.673$, each strictly above RecodAI-LUC's headline, and CoMoFoD further reaches AUPRC $0.833$, MCC $0.499$, and TPR $= 32.5\%$ at $1\%$ FPR.
Crucially, four structural claims of this paper are reproduced on every benchmark: the Laplacian-over-raw advantage (three of four datasets in matched comparison), the decoder-heavy layer ranking (\texttt{up\_blocks.2}/\texttt{up\_blocks.3} top-weighted in all four best configurations), the Wasserstein-dominant feature contribution, and the above-chance signal itself.
What varies is the fusion depth ($k \in \{2, 3, 5\}$) and the richness of the best feature bundle, both of which adapt to dataset scale in a predictable way.
The portable contribution is therefore a \emph{mechanism}---approximate subgraph duplication induces a detectable spectral redistribution in diffusion attention Laplacians---and not a single frozen hyperparameter recipe, which we view as a strength rather than a weakness for a training-free forensic framework.

\subsection{Honest Assessment of Signal Strength}
\label{sec:discussion:honest}

We provide a candid evaluation of the detector's practical utility:

\begin{itemize}
    \item \textbf{RecodAI-LUC headline AUROC of $0.606$ is statistically significant but operationally modest.} At $\le 1\%$ false positives on this benchmark, the detector catches $12.1\%$ of forgeries, and part of this ceiling reflects the benchmark's difficulty rather than the method itself, as evidenced by the substantially higher low-FPR TPRs on CoMoFoD (\Cref{sec:results:external}).
    \item \textbf{The approach is training-free}---no forgery labels are used at any stage, neither for calibration, layer selection, nor threshold setting. The appropriate comparison class is other zero-shot or unsupervised baselines, not supervised CMF detectors that fit to labelled forgery data.
    \item \textbf{The signal is robust across configurations and datasets.} On RecodAI-LUC, all Laplacian settings yield AUROC $\in [0.595, 0.606]$; across the four benchmarks evaluated (\Cref{sec:results:external}), AUROC spans $0.606$--$0.774$ with each dataset's best configuration using a Laplacian-based feature set and a decoder-heavy top-$k$ fusion. The signal is therefore not an artefact of a specific hyperparameter choice or a specific benchmark.
    \item \textbf{Cross-dataset transfer is multi-benchmark, not single-dataset.} MICC-F220, CoMoFoD, and COVERAGE all produce above-chance detection, and CoMoFoD reaches deployment-relevant operating points (TPR $= 32.5\%$ at $1\%$ FPR, MCC $= 0.499$). Bootstrap confidence intervals on the smaller benchmarks are correspondingly wider than on RecodAI-LUC, and this is stated explicitly in \Cref{tab:external_validation} and the limitations section.
    \item \textbf{The contribution is a principled, multi-benchmark-validated graph-spectral framework} connecting forgery theory to detection practice, not a new state-of-the-art number against supervised detectors trained end-to-end on labelled CMF data.
\end{itemize}

\subsection{Failure Modes and Negative Results}
\label{sec:discussion:failures}

\paragraph{Duplication mass did not validate the simple hypothesis.}
The original hypothesis that copy-move forgery should increase eigenvalue mass near $\lambda = 1$ was not confirmed. Instead, forged images show \emph{lower} near-one mass, suggesting that the perturbation disrupts rather than reinforces the duplication-band structure. This negative result guided our shift toward global transport features.

\paragraph{Causal weighting hurts detection.}
While useful for interpretation, incorporating causal drop ratios into fusion weights slightly degrades detection ($-0.006$ AUROC). The layers with the highest causal sensitivity are not necessarily those with the best image-level discriminability.

\paragraph{Localisation was not achieved.}
Mask-based localisation metrics (IoU, Dice, AUROC$_{\text{loc}}$) yielded near-zero scores across all layers, indicating that the spectral approach in its current form does not support pixel-level forgery localisation. This is expected: our method operates on global graph eigenvalues, which discard spatial information.

\section{Limitations and Future Work}
\label{sec:limitations}

\paragraph{Moderate detection accuracy.}
The best AUROC ($0.606$) falls below strongly supervised copy-move detectors that are trained end-to-end on labelled forgery data---including BusterNet~\cite{wu2018busternet}, DOA-GAN~\cite{islam2020doa}, and serial-localisation pipelines such as Chen et al.~\cite{chen2021serial}---which frequently report AUROC above $0.9$ on standard benchmarks. This gap is expected: our detector is training-free and uses no forgery labels at any stage, so the appropriate comparison class is unsupervised and zero-shot forensic methods rather than fully supervised classifiers.
Combining spectral features with learned representations, or using the spectral score as an auxiliary channel in a multi-modal detector, could improve performance.

\paragraph{Asymmetric evaluation scale across benchmarks.}
Headline AUROCs, ablations, and falsification tests are reported on four benchmarks, but RecodAI-LUC ($n_{\text{val}} = 1{,}282$) supports substantially more detailed analysis---per-layer tables, causal perturbation, and specificity experiments---than the three additional benchmarks, whose evaluation sets are smaller ($n \in \{40, 44, 80\}$) and therefore carry wider bootstrap confidence intervals (\Cref{tab:external_validation}).
Although the core conclusions (Laplacian advantage, decoder-layer dominance, Wasserstein-dominant signal) reproduce on all four datasets, the cross-dataset numbers should still be read as multi-benchmark validation of a mechanism rather than as head-to-head leaderboard results under a single harmonised protocol.
A stronger leaderboard-style claim would require repeated multi-dataset experiments with fixed preprocessing, fixed split rules, and uniformly reported uncertainty across all datasets, which we view as a natural next step.

\paragraph{Single diffusion backbone.}
All experiments use SD v1.5.
Testing with SD 2.1, SDXL, or alternative architectures would assess whether the spectral forensic signal generalises across model families.

\paragraph{Falsification experiments completed.}
All four specificity experiments were executed (\Cref{sec:results:falsification}). While results confirm monotonic sensitivity and partial copy-move specificity, the natural self-similarity confound (high-repetition authentic images scoring moderately) and non-zero scores for generic corruptions suggest the spectral signal is not purely copy-move-specific.

\paragraph{Computational cost.}
Full eigendecomposition at high-resolution layers ($64 \times 64 = 4{,}096$ tokens) is costly.
Approximate spectral methods (Lanczos, randomised SVD) or spectral moment estimation could reduce runtime.

\paragraph{Theoretical gap.}
Our perturbation-theoretic arguments are qualitative.
A rigorous bound relating copy-move region size to minimum detectable Wasserstein shift would strengthen the theoretical foundation.

\section{Conclusion}
\label{sec:conclusion}

We introduced GraphSpecForge, a training-free framework for copy-move forgery detection based on spectral analysis of diffusion attention graphs.
By connecting copy-move manipulation to approximate subgraph duplication and its spectral consequences through normalized graph Laplacians, we established a principled forensic paradigm distinct from both classical feature matching and supervised deep learning approaches.

Our experiments on the large-scale RecodAI-LUC benchmark and three additional public copy-move datasets (MICC-F220, CoMoFoD, COVERAGE) demonstrate that:
(1) the normalized Laplacian spectrum consistently outperforms raw attention spectra on three of four benchmarks under matched settings, and across the full ablation envelope;
(2) Wasserstein transport distance carries most of the forensic signal on every benchmark;
(3) decoder layers at mid-resolution---\texttt{up\_blocks.2} and \texttt{up\_blocks.3}---are the top-ranked FSEL layers on all four datasets;
(4) the detection signal is statistically significant on RecodAI-LUC ($p = 0.005$) and reaches deployment-relevant AUROCs of $0.752$--$0.774$ on MICC-F220 and CoMoFoD, with CoMoFoD attaining MCC $= 0.499$ and TPR $= 32.5\%$ at $1\%$ FPR;
(5) falsification experiments confirm partial copy-move specificity, with monotonically increasing sensitivity to manipulation strength, collapse under null-graph label shuffling, and higher scores for genuine forgeries than non-copy-move corruptions;
(6) non-spectral graph controls and filter-bank features provide modest complementary benefit when combined with Wasserstein transport features; and
(7) the cross-dataset evaluation establishes that the spectral anomaly \emph{mechanism} is portable across copy-move curation protocols, while the optimal feature bundle and fusion depth calibrate to dataset scale in a predictable way.

The primary contribution of this work is not state-of-the-art accuracy, but rather a theoretically grounded analytical framework that opens new directions for leveraging the rich structural representations in diffusion models for image forensics.
Future work combining graph-spectral features with learned detectors, extending to multiple manipulation types, and establishing formal detection guarantees could bring this paradigm closer to practical deployment.

\bibliographystyle{plain}

\newpage
\appendix
\section{Reproducibility}
\label{app:reproducibility}

\paragraph{Software.}
Python 3.10, PyTorch 2.1, diffusers 0.24, NumPy 1.24, SciPy 1.11, scikit-learn 1.3, matplotlib 3.8.

\paragraph{Hardware.}
All experiments on a single NVIDIA H100 (80~GB VRAM) via Kaggle.
Full pipeline runtime: ${\sim}4$ hours for 5{,}128 images across 16 layers.
Extended ablation on 400-image subset: ${\sim}3$ minutes.

\paragraph{Hyperparameters.}
Tail quantile $q = 0.10$.
Bootstrap resamples $B = 200$.
Near-one epsilon $\varepsilon = 0.05$.
FSEL weights: $w_d = 0.45$, $w_l = 0.25$, $w_c = 0.20$, $w_s = 0.10$.
Top-$k$: $k = 5$ (selected by validation).
Temperature $\tau = 1.0$.

\paragraph{Data availability.}
The RecodAI-LUC CMF Dataset is available on Kaggle.
Code and result CSVs will be released upon publication.

Extended results including full per-layer metrics, reliability decompositions, $\alpha_{\text{Hill}}$ analysis, causal perturbation plots, FSEL heatmaps, and additional diagnostic figures are provided in the \textbf{Supplementary Material}.

\end{document}


\maketitle

This document provides extended results, additional figures, and detailed per-layer and per-dataset metrics that complement the main paper.
Section~\ref{supp:esd}--\ref{supp:diagnostics} report extended results on the primary RecodAI-LUC benchmark;
Section~\ref{supp:pairwise} adds a qualitative one-vs-one case study;
Sections~\ref{supp:external}--\ref{supp:external_layers} report the full external-dataset ablations and per-layer rankings underlying the pilot transfer experiments in the main paper.

\section{Empirical Spectral Density Comparison}
\label{supp:esd}

\begin{figure}[!htbp]
\centering
\includegraphics[width=\linewidth,height=0.82\textheight,keepaspectratio]{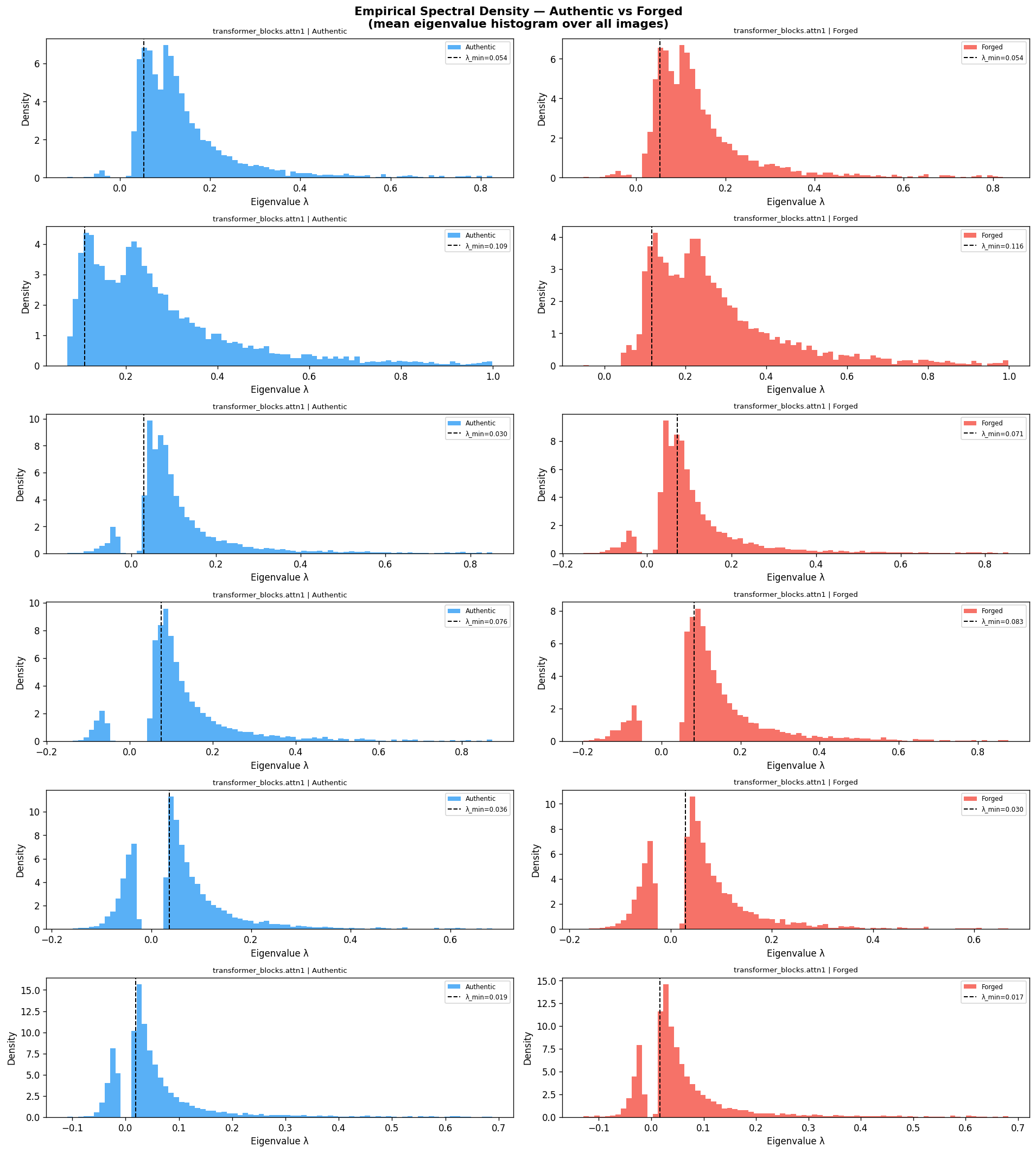}
\caption{\textbf{Empirical spectral density (ESD) of raw attention eigenvalues across layers.} Left column: authentic images (blue). Right column: forged images (red). Each row corresponds to a different U-Net attention layer. Subtle but systematic shifts in the spectral distribution are visible, particularly in the tail region, supporting the hypothesis that copy-move forgery induces spectral redistribution.}
\label{fig:supp_esd_comparison}
\end{figure}

\FloatBarrier
\section{Extended Layer-Wise Wasserstein Results}
\label{supp:layers}

\begin{table}[!htbp]
\centering
\caption{\textbf{Complete per-layer Wasserstein metrics on RecodAI-LUC.} $W_1^F$: full Wasserstein ($\times 10^{-3}$). $W_1^T$: tail Wasserstein ($\times 10^{-2}$). KS: Kolmogorov--Smirnov statistic ($\times 10^{-2}$). $d$: Cohen's $d$ on image-level means.}
\label{tab:supp_all_layers}
\small
\setlength{\tabcolsep}{3pt}
\begin{tabular}{@{}lrrrrrl@{}}
\toprule
\textbf{Layer} & $\boldsymbol{W_1^F}$ & $\boldsymbol{W_1^T}$ & \textbf{KS} & $\boldsymbol{d}$ & \textbf{AUROC} & \textbf{CI}  \\
\midrule
\texttt{up3.attn0}  & 5.504 & 3.518 & 2.207 & 0.306 & 0.526 & [5.44, 5.57] \\
\texttt{up2.attn0}  & 4.713 & 2.014 & 1.844 & 0.201 & 0.540 & [4.57, 4.86] \\
\texttt{up2.attn1}  & 4.197 & 2.014 & 2.159 & 0.237 & 0.485 & [4.07, 4.33] \\
\texttt{up1.attn1}  & 4.067 & 1.070 & 1.591 & 0.179 & 0.567 & [3.77, 4.46] \\
\texttt{up3.attn1}  & 3.981 & 1.634 & 2.431 & 0.220 & 0.524 & [3.92, 4.04] \\
\texttt{up3.attn2}  & 2.433 & 1.077 & 1.776 & 0.177 & 0.531 & [2.39, 2.48] \\
\texttt{up1.attn0}  & 1.483 & 0.639 & 1.235 & 0.201 & 0.531 & [1.31, 1.78] \\
\texttt{mid.attn0}  & 1.422 & 0.817 & 1.366 & 0.243 & 0.516 & [1.28, 2.26] \\
\texttt{down2.attn1} & 1.264 & 0.404 & 0.804 & 0.077 & 0.505 & [1.15, 1.45] \\
\texttt{down1.attn1} & 0.983 & 0.318 & 0.495 & 0.025 & 0.496 & [0.92, 1.08] \\
\texttt{down2.attn0} & 0.940 & 0.312 & 0.509 & 0.057 & 0.489 & [0.81, 1.16] \\
\texttt{up1.attn2}  & 0.824 & 0.404 & 1.325 & 0.112 & 0.457 & [0.68, 1.13] \\
\texttt{up2.attn2}  & 0.803 & 0.422 & 1.226 & 0.117 & 0.471 & [0.74, 0.91] \\
\texttt{down1.attn0} & 0.428 & 0.161 & 0.508 & 0.027 & 0.476 & [0.37, 0.51] \\
\texttt{down0.attn1} & 0.367 & 0.128 & 0.452 & 0.022 & 0.485 & [0.34, 0.41] \\
\texttt{down0.attn0} & 0.174 & 0.072 & 0.759 & 0.015 & 0.488 & [0.16, 0.20] \\
\bottomrule
\end{tabular}
\end{table}

\FloatBarrier
\section{Extended Laplacian Detection Metrics}
\label{supp:laplacian}

\begin{table}[!htbp]
\centering
\caption{\textbf{Per-layer validation AUROC for Laplacian and raw spectra on RecodAI-LUC.} Sorted by Laplacian AUROC. Top-3 in bold.}
\label{tab:supp_all_laplacian}
\small
\begin{tabular}{@{}lcc@{}}
\toprule
\textbf{Layer} & \textbf{Laplacian AUROC} & \textbf{Raw AUROC} \\
\midrule
\texttt{up2.attn0}  & \textbf{0.603} & 0.540 \\
\texttt{up2.attn1}  & \textbf{0.597} & 0.550 \\
\texttt{up3.attn0}  & \textbf{0.595} & 0.515 \\
\texttt{up1.attn1}  & 0.583 & 0.603 \\
\texttt{up3.attn1}  & 0.576 & 0.523 \\
\texttt{up3.attn2}  & 0.561 & 0.525 \\
\texttt{up1.attn0}  & 0.546 & 0.564 \\
\texttt{down1.attn1} & 0.541 & 0.512 \\
\texttt{mid.attn0}  & 0.532 & 0.533 \\
\texttt{down2.attn1} & 0.529 & 0.543 \\
\texttt{up2.attn2}  & 0.521 & 0.514 \\
\texttt{down2.attn0} & 0.511 & 0.502 \\
\texttt{down1.attn0} & 0.508 & 0.497 \\
\texttt{down0.attn1} & 0.502 & 0.471 \\
\texttt{up1.attn2}  & 0.502 & 0.537 \\
\texttt{down0.attn0} & 0.484 & 0.487 \\
\bottomrule
\end{tabular}
\end{table}

\FloatBarrier
\section{Reliability and FSEL Details}
\label{supp:fsel}

The FSEL score combines four normalised components.
Table~\ref{tab:supp_reliability} shows the reliability decomposition for the top layers under the Laplacian spectrum.

\begin{table}[!htbp]
\centering
\caption{\textbf{Reliability decomposition for top Laplacian layers.} $z_A$: normalised AUROC. $z_S$: normalised separation. $z_C$: normalised causal drop. $z_{\text{CI}}$: CI width penalty. $R_{\text{no-c}}$: reliability without causal. $R_c$: reliability with causal.}
\label{tab:supp_reliability}
\small
\setlength{\tabcolsep}{3pt}
\begin{tabular}{@{}lrrrrrr@{}}
\toprule
\textbf{Layer} & $\boldsymbol{z_A}$ & $\boldsymbol{z_S}$ & $\boldsymbol{z_C}$ & $\boldsymbol{z_{\text{CI}}}$ & $\boldsymbol{R_{\text{no-c}}}$ & $\boldsymbol{R_c}$ \\
\midrule
\texttt{up2.attn0}  &  1.93 &  1.97 & $-0.39$ & $-0.03$ &  2.62 &  2.31 \\
\texttt{up2.attn1}  &  1.33 &  2.07 & $-0.12$ & $-0.15$ &  2.33 &  2.24 \\
\texttt{up3.attn0}  &  1.43 &  1.58 &  1.62  & $-0.71$ &  2.48 &  3.77 \\
\texttt{up1.attn1}  &  1.00 &  0.43 & $-2.01$ &  1.62  & $-0.15$ & $-1.76$ \\
\texttt{up3.attn1}  &  0.55 & $-0.10$ & $-0.44$ & $-0.71$ &  0.34 & $-0.01$ \\
\bottomrule
\end{tabular}
\end{table}

\begin{figure}[!htbp]
\centering
\begin{subfigure}[b]{\linewidth}
    \centering
    \includegraphics[width=\linewidth]{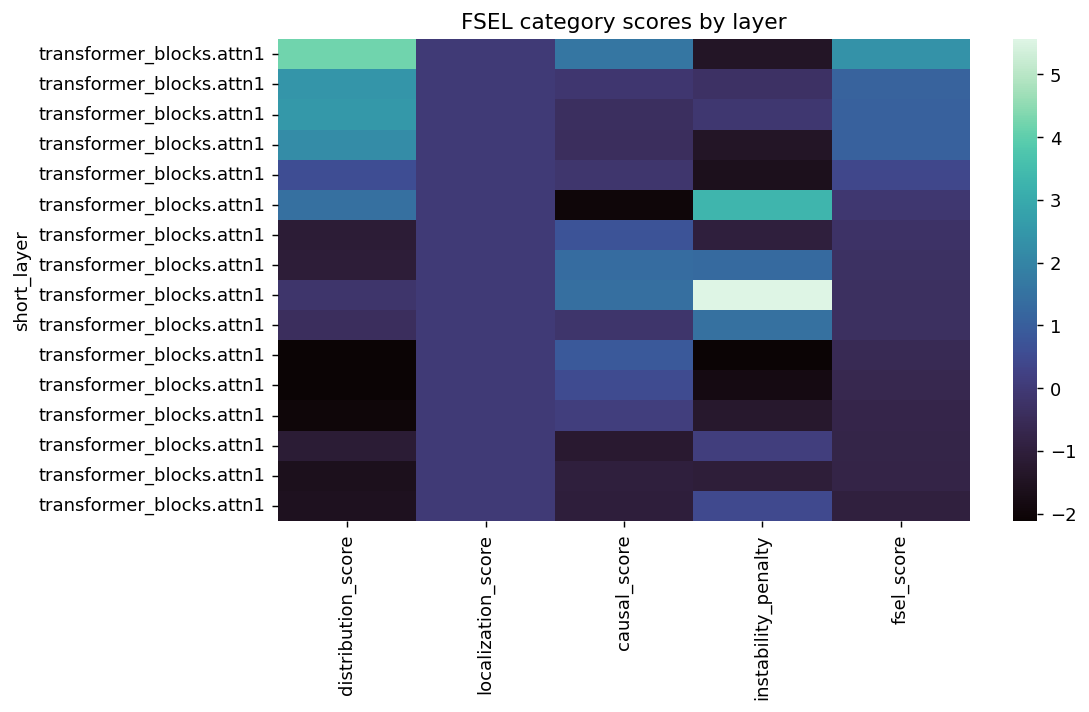}
    \caption{FSEL component scores by layer}
\end{subfigure}

\vspace{0.6em}

\begin{subfigure}[b]{\linewidth}
    \centering
    \includegraphics[width=\linewidth]{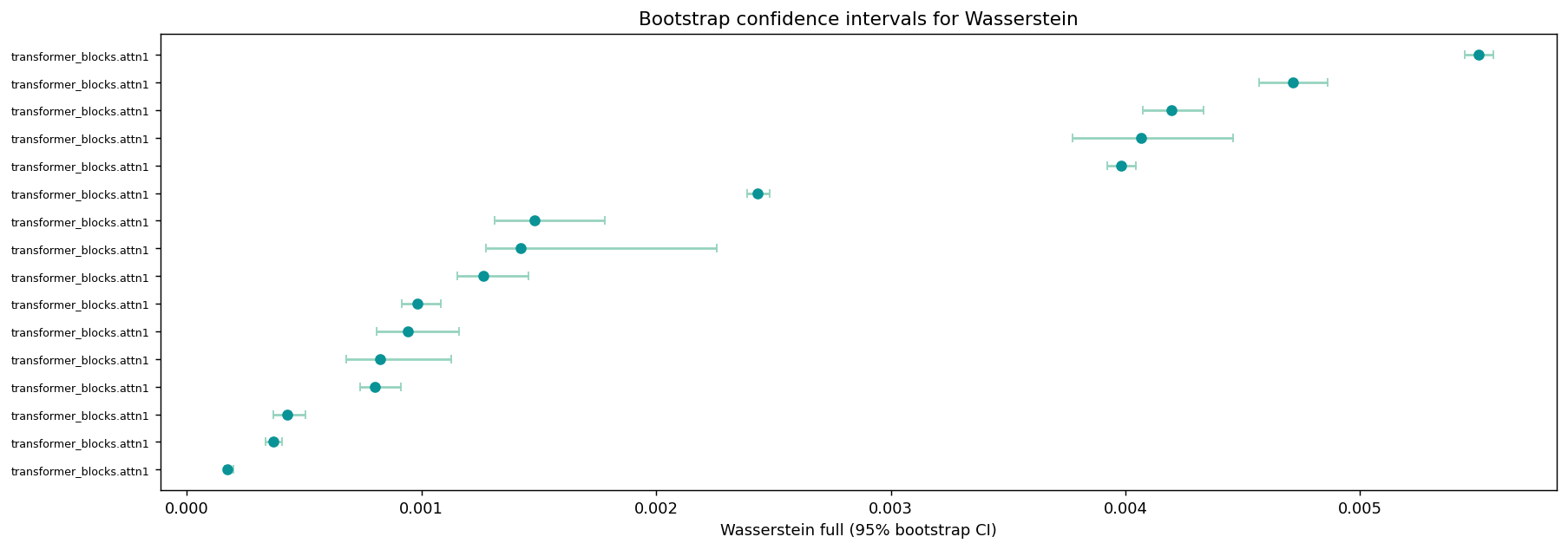}
    \caption{Bootstrap 95\% CIs for pooled $W_1$ per layer}
\end{subfigure}
\caption{\textbf{FSEL decomposition and statistical precision.} (a) Heatmap of FSEL component scores (distribution, localisation, causal, instability, composite) across all 16 layers. (b) Bootstrap 95\% confidence intervals for the pooled Wasserstein distance at each layer.}
\label{fig:supp_fsel_heatmap_ci}
\end{figure}

\paragraph{FSEL heatmap and bootstrap confidence intervals.}
\Cref{fig:supp_fsel_heatmap_ci} visualises the same decomposition across all 16 layers and shows bootstrap 95\% confidence intervals on the pooled Wasserstein distance per layer.
\label{supp:fsel_heatmap}

\FloatBarrier
\section{Laplacian Detector Layer-Level Separation}
\label{supp:laplacian_sep}

Table~\ref{tab:supp_laplacian_sep} consolidates four per-layer spectral separation metrics computed on the 400-image FSEL subset.
$W_1^F$ is the pooled full-spectrum Laplacian Wasserstein distance; $\Delta\bar{W}_1$ is the mean image-level score gap (forged $-$ authentic); AUROC is the single-layer validation AUROC under the Laplacian spectrum (cf.\ Table~\ref{tab:supp_all_laplacian}); $\delta_c$ is the causal drop ratio (cf.\ Table~\ref{tab:supp_causal}).
Decoder layers (\texttt{up\_blocks}) dominate on all four metrics: \texttt{up3.attn0} has the largest $W_1^F$ and $\Delta\bar{W}_1$, while \texttt{up2.attn0/1} lead on AUROC.
Encoder layers (\texttt{down\_blocks}) show the smallest separation on every metric, confirming that the forgery-induced spectral redistribution is concentrated in the high-resolution decoder stages of the U-Net.

\begin{table}[!htbp]
\centering
\caption{\textbf{Per-layer Laplacian spectral separation (400-image FSEL subset).}
Sorted by $W_1^F$ descending.
$W_1^F$: pooled full-spectrum Wasserstein ($\times 10^{-3}$).
$\Delta\bar{W}_1$: mean image-level score gap, forged$-$authentic ($\times 10^{-3}$).
AUROC: validation AUROC (Laplacian spectrum).
$\delta_c$: causal drop ratio.}
\label{tab:supp_laplacian_sep}
\small
\setlength{\tabcolsep}{5pt}
\begin{tabular}{@{}lrrrr@{}}
\toprule
\textbf{Layer} & $\boldsymbol{W_1^F}$ & $\boldsymbol{\Delta\bar{W}_1}$ & \textbf{AUROC} & $\boldsymbol{\delta_c}$ \\
\midrule
\texttt{up3.attn0}   & 5.335 & 4.95 & 0.595 & 0.481 \\
\texttt{up2.attn0}   & 4.562 & 2.17 & 0.603 & 0.294 \\
\texttt{up2.attn1}   & 4.040 & 2.95 & 0.597 & 0.319 \\
\texttt{up1.attn1}   & 3.995 & 2.07 & 0.583 & 0.151 \\
\texttt{up3.attn1}   & 3.644 & 3.64 & 0.576 & 0.287 \\
\texttt{up3.attn2}   & 2.072 & 2.07 & 0.561 & 0.342 \\
\texttt{down2.attn1} & 1.418 & 0.73 & 0.529 & 0.205 \\
\texttt{up1.attn0}   & 1.444 & 0.97 & 0.546 & 0.315 \\
\texttt{mid.attn0}   & 1.351 & 0.96 & 0.532 & 0.451 \\
\texttt{down2.attn0} & 1.095 & 0.54 & 0.511 & 0.186 \\
\texttt{down1.attn1} & 1.089 & 0.21 & 0.541 & 0.205 \\
\texttt{up1.attn2}   & 0.874 & 0.87 & 0.502 & 0.481 \\
\texttt{up2.attn2}   & 0.760 & 0.73 & 0.521 & 0.412 \\
\texttt{down0.attn1} & 0.652 & 0.51 & 0.502 & 0.256 \\
\texttt{down1.attn0} & 0.509 & 0.27 & 0.508 & 0.292 \\
\texttt{down0.attn0} & 0.311 & 0.16 & 0.484 & 0.292 \\
\bottomrule
\end{tabular}
\end{table}

\FloatBarrier
\section{Causal Perturbation and Layer Weights}
\label{supp:causal}

\paragraph{Localisation diagnostics.}
Pixel-level localisation metrics (IoU, Dice, AUROC$_{\text{loc}}$, AUPRC$_{\text{loc}}$) are exactly zero for every attention layer on the 400-image FSEL subset.
This is expected: the detector operates at the \emph{image level}; self-attention maps are aggregated globally and no mask supervision is provided, so the pipeline carries no per-pixel forgery localisation signal.

\paragraph{Softmax fusion weights.}
Table~\ref{tab:supp_fusion_weights} reports the per-layer reliability score (sum of AUROC and separation $z$-scores) and the corresponding softmax fusion weight for the top-$k=3$ layers under both Laplacian and raw-spectrum detectors.
The Laplacian top-three layers have closely matched reliability and closely matched softmax weights (0.373 / 0.325 / 0.302), indicating that the fusion is not over-reliant on any single layer.
The raw-spectrum top-three show a similar spread (0.375 / 0.339 / 0.286).
In both cases all three fused layers belong to the decoder (\texttt{up\_blocks}), consistent with the separation analysis in Table~\ref{tab:supp_laplacian_sep}.
The full per-layer causal drop analysis follows in Table~\ref{tab:supp_causal}.

\begin{table}[!htbp]
\centering
\caption{\textbf{Top-3 layer fusion weights ($k=3$, softmax-reliability).}
Reliability $= z_A + z_S$ (normalised AUROC $+$ normalised separation scores).
Softmax weight $= e^{R_\ell}\,/\!\sum_{j=1}^{3}e^{R_j}$.
AUROC is validation AUROC for the single-layer detector.}
\label{tab:supp_fusion_weights}
\small
\setlength{\tabcolsep}{4pt}
\begin{tabular}{@{}llrrr@{}}
\toprule
\textbf{Spectrum} & \textbf{Layer} & \textbf{AUROC (val)} & \textbf{Reliability} & \textbf{Softmax wt.} \\
\midrule
Laplacian & \texttt{up2.attn1} & 0.628 & 2.776 & 0.373 \\
Laplacian & \texttt{up3.attn0} & 0.597 & 2.642 & 0.325 \\
Laplacian & \texttt{up2.attn0} & 0.614 & 2.567 & 0.302 \\
\midrule
Raw       & \texttt{up1.attn1} & 0.606 & 1.979 & 0.375 \\
Raw       & \texttt{up2.attn1} & 0.590 & 1.878 & 0.339 \\
Raw       & \texttt{up1.attn0} & 0.608 & 1.709 & 0.286 \\
\bottomrule
\end{tabular}
\end{table}

\begin{table}[!htbp]
\centering
\caption{\textbf{Causal perturbation metrics by layer} (400-image FSEL subset). $W_1^{\mathrm{pool}}$: pooled authentic--forged Wasserstein ($\times 10^{-3}$). $\delta_c$: causal drop ratio. $W_1^{\mathrm{ablate}}$: implied pooled distance after ablating the top eigencomponents, $(1-\delta_c)\, W_1^{\mathrm{pool}}$.}
\label{tab:supp_causal}
\small
\setlength{\tabcolsep}{4pt}
\begin{tabular}{@{}lccc@{}}
\toprule
\textbf{Layer} & $\boldsymbol{W_1^{\mathrm{pool}}}$ & $\boldsymbol{\delta_c}$ & $\boldsymbol{W_1^{\mathrm{ablate}}}$ \\
\midrule
\texttt{up3.attn0}  & 5.335 & 0.481 & 2.769 \\
\texttt{up2.attn1}  & 4.040 & 0.319 & 2.753 \\
\texttt{up2.attn0}  & 4.562 & 0.294 & 3.222 \\
\texttt{up3.attn1}  & 3.644 & 0.287 & 2.596 \\
\texttt{up3.attn2}  & 2.072 & 0.342 & 1.363 \\
\texttt{up1.attn1}  & 3.995 & 0.151 & 3.391 \\
\texttt{up1.attn2}  & 0.874 & 0.481 & 0.454 \\
\texttt{up2.attn2}  & 0.760 & 0.412 & 0.447 \\
\texttt{up1.attn0}  & 1.444 & 0.315 & 0.988 \\
\texttt{mid.attn0}  & 1.351 & 0.451 & 0.741 \\
\texttt{down2.attn1} & 1.418 & 0.205 & 1.127 \\
\texttt{down0.attn0} & 0.311 & 0.292 & 0.220 \\
\texttt{down0.attn1} & 0.652 & 0.256 & 0.485 \\
\texttt{down1.attn0} & 0.509 & 0.292 & 0.360 \\
\texttt{down1.attn1} & 1.089 & 0.205 & 0.866 \\
\texttt{down2.attn0} & 1.095 & 0.186 & 0.891 \\
\bottomrule
\end{tabular}
\end{table}

\FloatBarrier
\section{Alpha-Hill Tail Analysis}
\label{supp:alpha_hill}

\begin{figure}[!htbp]
\centering
\includegraphics[width=\linewidth]{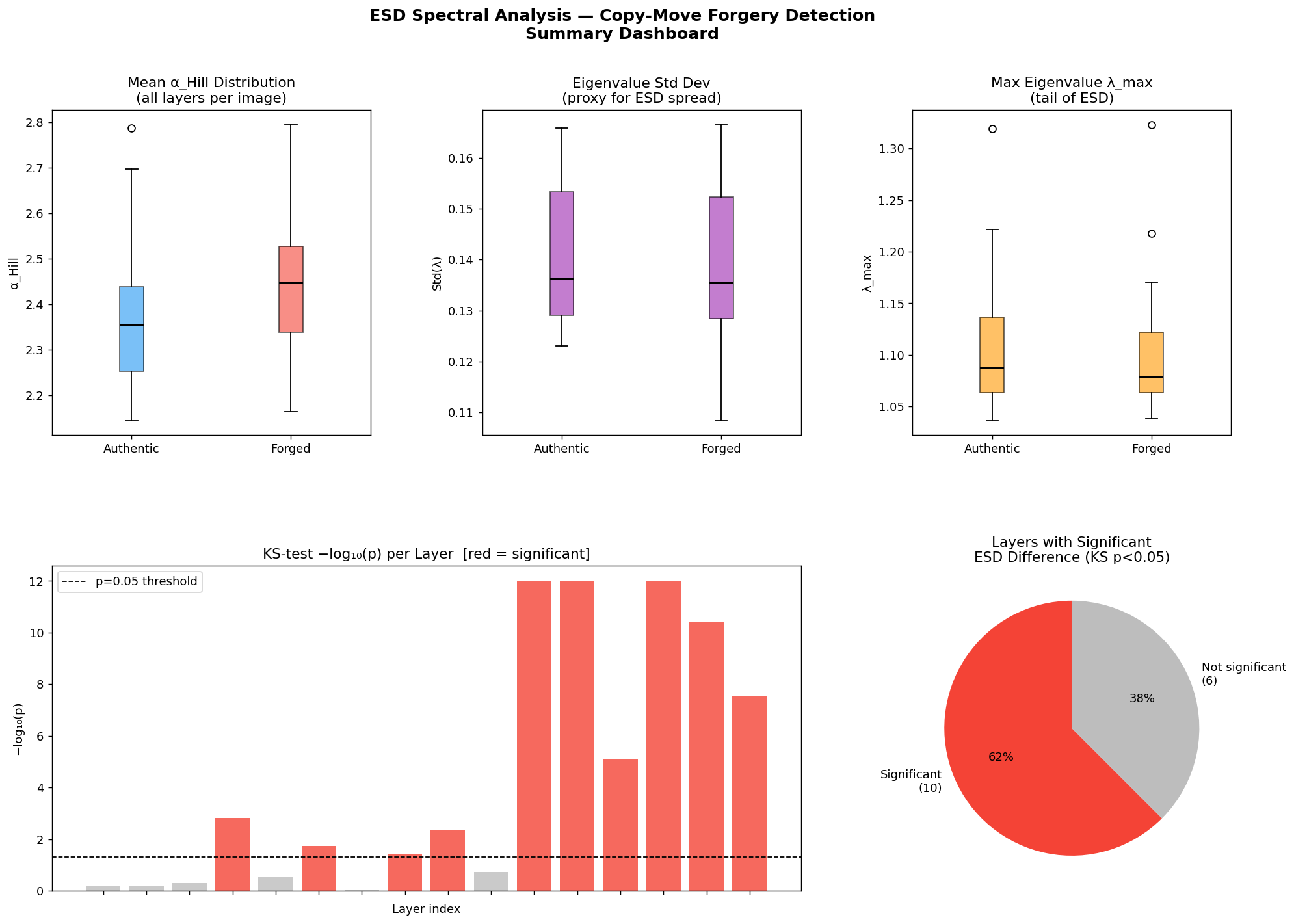}
\caption{\textbf{ESD spectral analysis summary dashboard.} Top row: distribution comparisons of mean $\alpha_{\text{Hill}}$, eigenvalue standard deviation, and maximum eigenvalue between authentic and forged images. Bottom left: per-layer KS test significance ($-\log_{10} p$). Bottom right: 62\% of layers show statistically significant ESD differences.}
\label{fig:supp_summary_dashboard}
\end{figure}

\begin{figure}[!htbp]
\centering
\includegraphics[width=\linewidth]{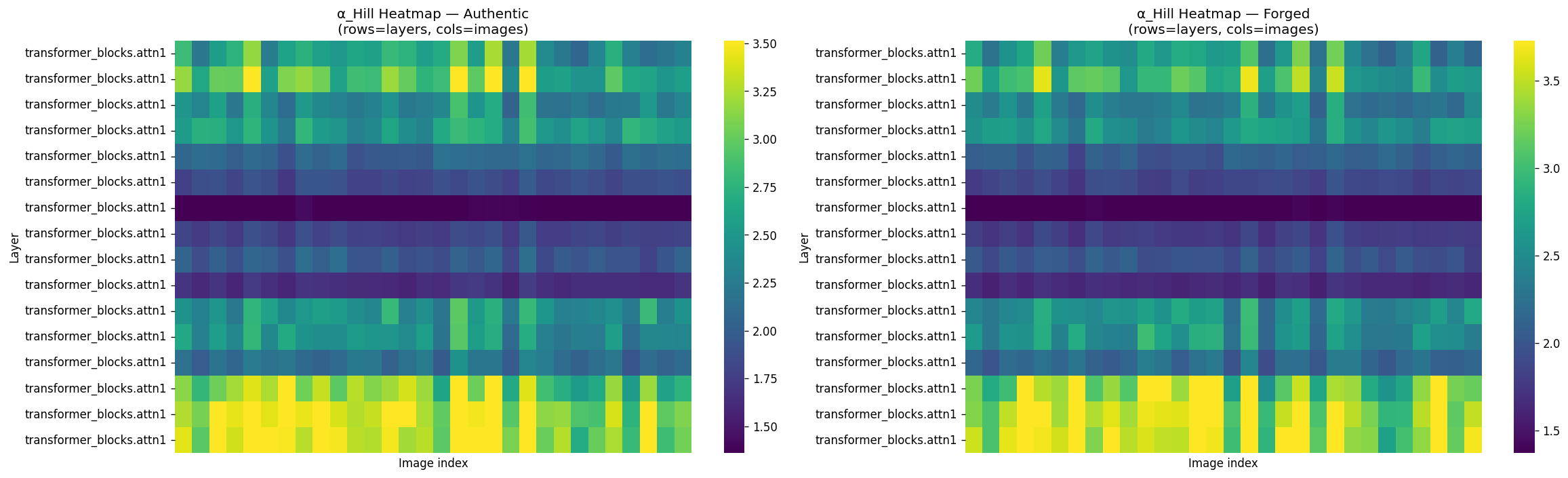}
\caption{\textbf{$\alpha_{\text{Hill}}$ heatmaps across layers and images.} Left: authentic images. Right: forged images. Each row is a U-Net attention layer; each column is an image.}
\label{fig:supp_alpha_hill_heatmap}
\end{figure}

\paragraph{Per-layer $\alpha_{\text{Hill}}$ trend.}
Forged images consistently exhibit higher $\alpha_{\text{Hill}}$ values (lighter spectral tails) than authentic images across all 16 layers.
The forged-minus-authentic mean difference $\Delta\alpha_{\text{Hill}} > 0$ at 15 of 16 layers; the pattern is visible in the heatmap above (Figure~\ref{fig:supp_alpha_hill_heatmap}) and in the distribution comparisons in Figure~\ref{fig:supp_summary_dashboard} (top-left panel).
This directional consistency corroborates the Wasserstein-based detector: copy-move manipulation disperses spectral energy into lighter tails, which the Laplacian normalisation amplifies into a measurable transport cost near $\lambda = 1$.

\FloatBarrier
\section{Additional Diagnostic Figures}
\label{supp:diagnostics}

\begin{figure}[!htbp]
\centering
\includegraphics[width=0.9\textwidth,keepaspectratio]{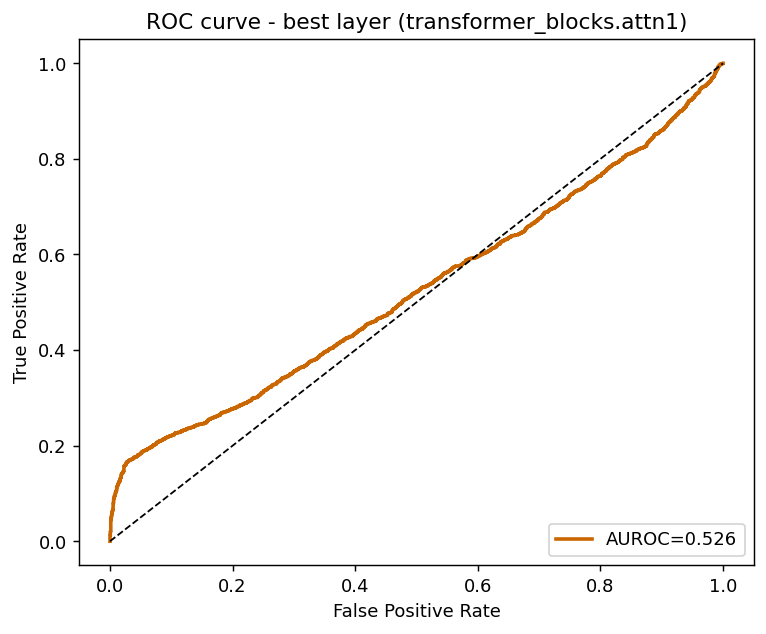}
\caption{\textbf{ROC curve for the best single layer} (raw attention spectrum, \texttt{up\_blocks.1.attn1}, AUROC = 0.567). Comparison with the fused Laplacian detector (AUROC = 0.606) shows the cumulative benefit of Laplacian normalisation and multi-layer fusion.}
\label{fig:supp_roc_single_raw}
\end{figure}

\begin{figure}[!htbp]
\centering
\includegraphics[width=\linewidth]{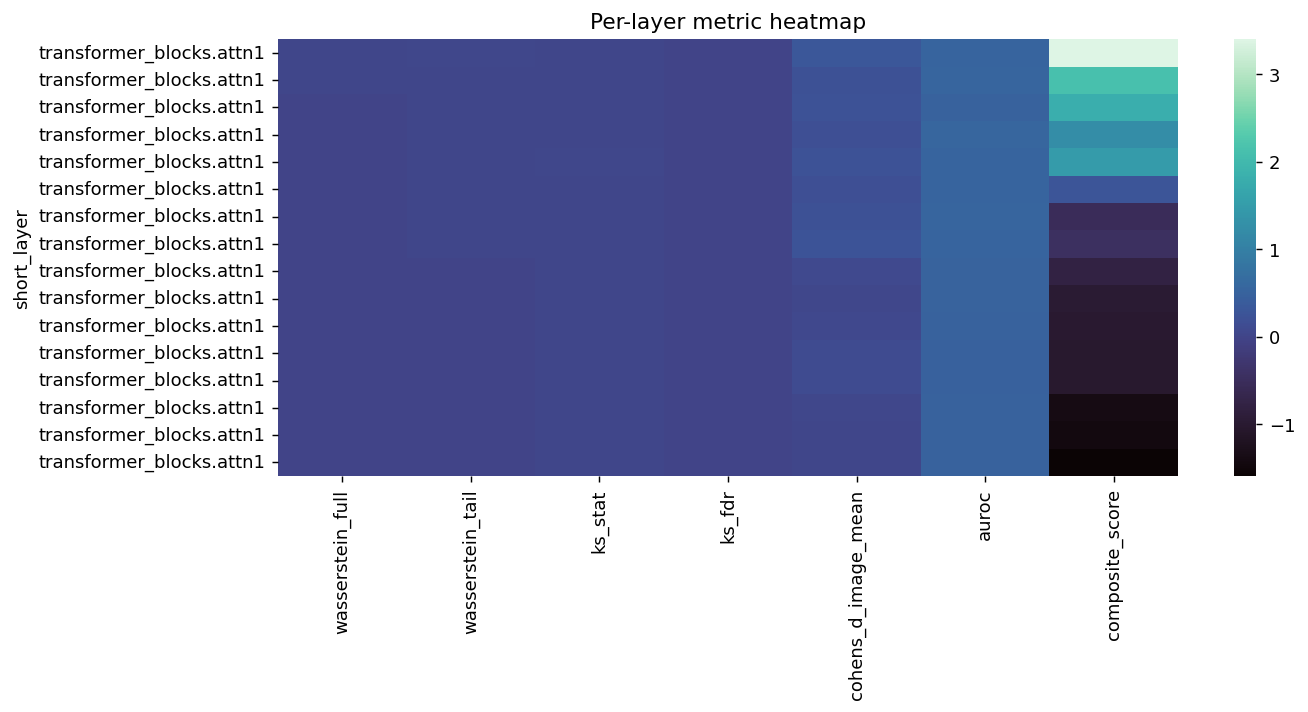}
\caption{\textbf{Per-layer metric heatmap.} Normalised metrics (Wasserstein full/tail, KS, Cohen's $d$, AUROC, composite score) across all 16 attention layers. Decoder layers consistently rank highest.}
\label{fig:supp_heatmap}
\end{figure}

\FloatBarrier
\section{Qualitative One-vs-One Pairwise Case Study}
\label{supp:pairwise}

To complement the aggregate statistics reported in the main paper, we examine a single authentic/forged pair from the RecodAI-LUC benchmark (pair \texttt{013}, files \texttt{013\_O.png} / \texttt{013\_F.png}) in isolation.
The pair is drawn from an \emph{a priori} pairwise analysis: we fix the attention layer to the top-ranked decoder layer, \texttt{up\_blocks.2.attentions.0.transformer\_blocks.0.attn1}, and compare the authentic and forged ESDs computed from \emph{only that image}, with $n = 1{,}024$ eigenvalues per spectrum.
This is a purely illustrative, per-image sanity check: no thresholds or detector parameters are tuned on it, and the main-benchmark AUROC is unchanged by its inclusion.

\paragraph{Pairwise Wasserstein shift.}
On this pair, the full-spectrum Laplacian Wasserstein distance is $\Wdist^{F} = 1.405 \times 10^{-2}$ and the tail Wasserstein is $\Wdist^{T} = 1.746 \times 10^{-2}$.
Under the raw attention spectrum at the same layer, $\Wdist^{F} = 1.379 \times 10^{-2}$ and $\Wdist^{T} = 3.388 \times 10^{-2}$.
The forged image's Laplacian ESD has systematically lower peak mass near $\lambda = 1$ ($0.534$ vs.\ $0.611$ for the authentic) and higher variance ($1.57 \times 10^{-2}$ vs.\ $1.28 \times 10^{-2}$), consistent with the \emph{redistribution-away-from-the-duplication-band} pattern reported at the population level in the main paper.

\begin{figure}[!htbp]
\centering
\begin{subfigure}[t]{0.48\linewidth}
    \centering
    \includegraphics[width=\linewidth]{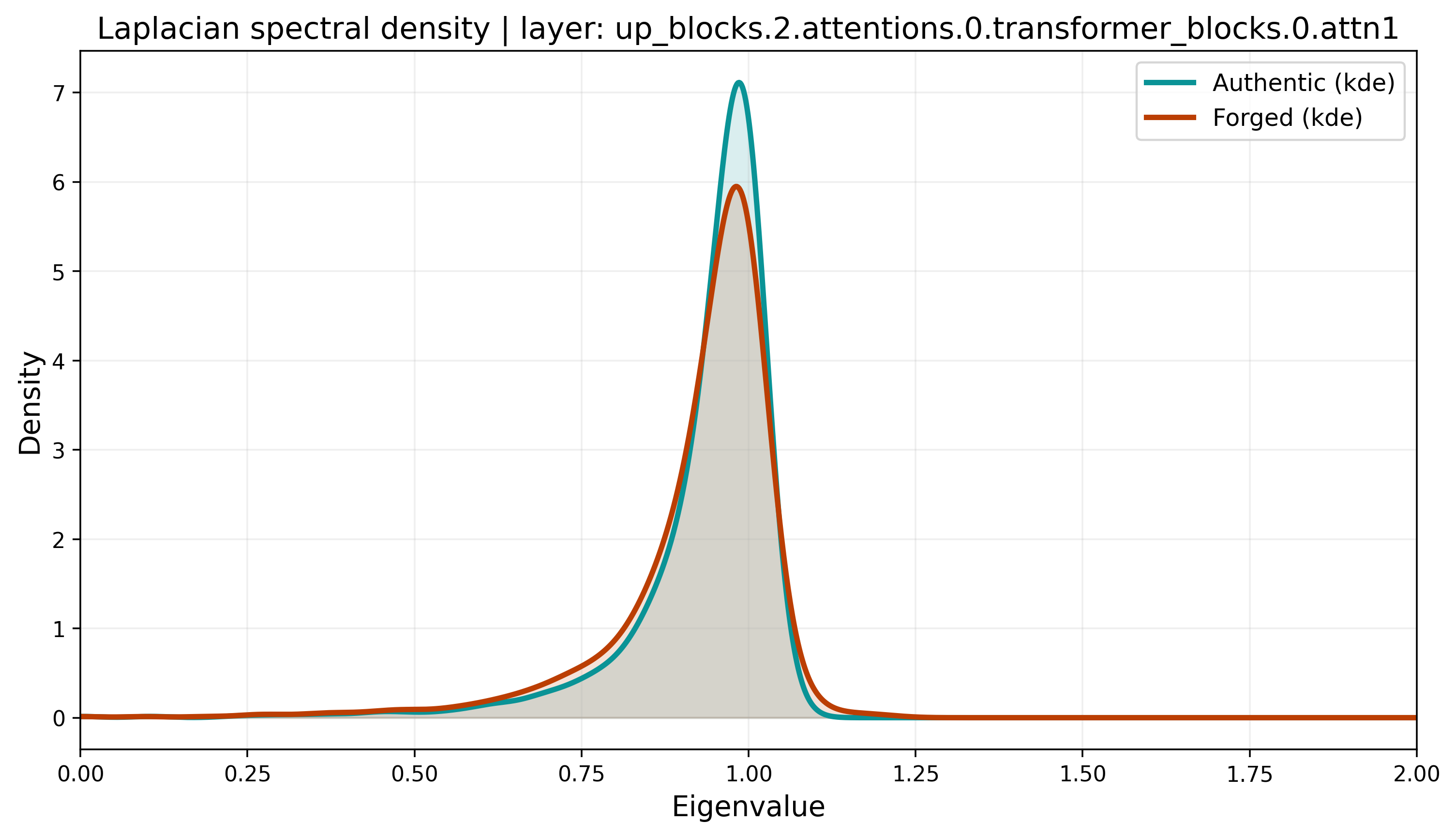}
    \caption{Laplacian density (authentic vs.\ forged)}
\end{subfigure}
\hfill
\begin{subfigure}[t]{0.48\linewidth}
    \centering
    \includegraphics[width=\linewidth]{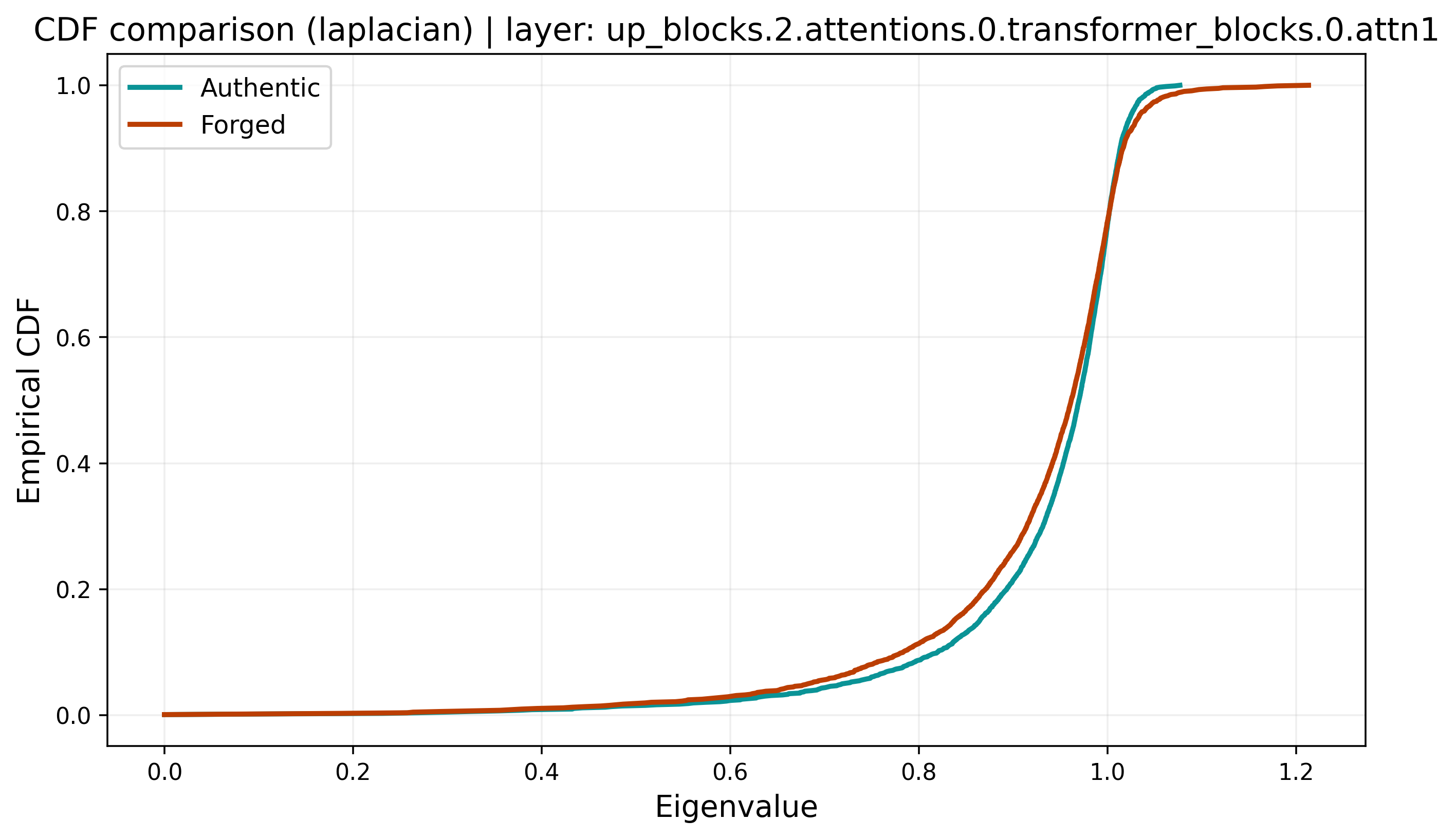}
    \caption{Laplacian CDF (authentic vs.\ forged)}
\end{subfigure}
\caption{\textbf{Per-image Laplacian ESD on pair \texttt{013}.}
(a) Kernel-density estimates of the per-image Laplacian spectra at \texttt{up\_blocks.2.attn0}; the forged spectrum is flatter around $\lambda = 1$ and heavier in the $[0.6, 0.9]$ and $[1.0, 1.2]$ bands.
(b) Corresponding empirical CDFs---the horizontal gap between the two curves near $\lambda \approx 1$ is the transport cost that drives the image-level Wasserstein score.}
\label{fig:supp_pairwise_laplacian}
\end{figure}

\begin{figure}[!htbp]
\centering
\begin{subfigure}[t]{0.48\linewidth}
    \centering
    \includegraphics[width=\linewidth]{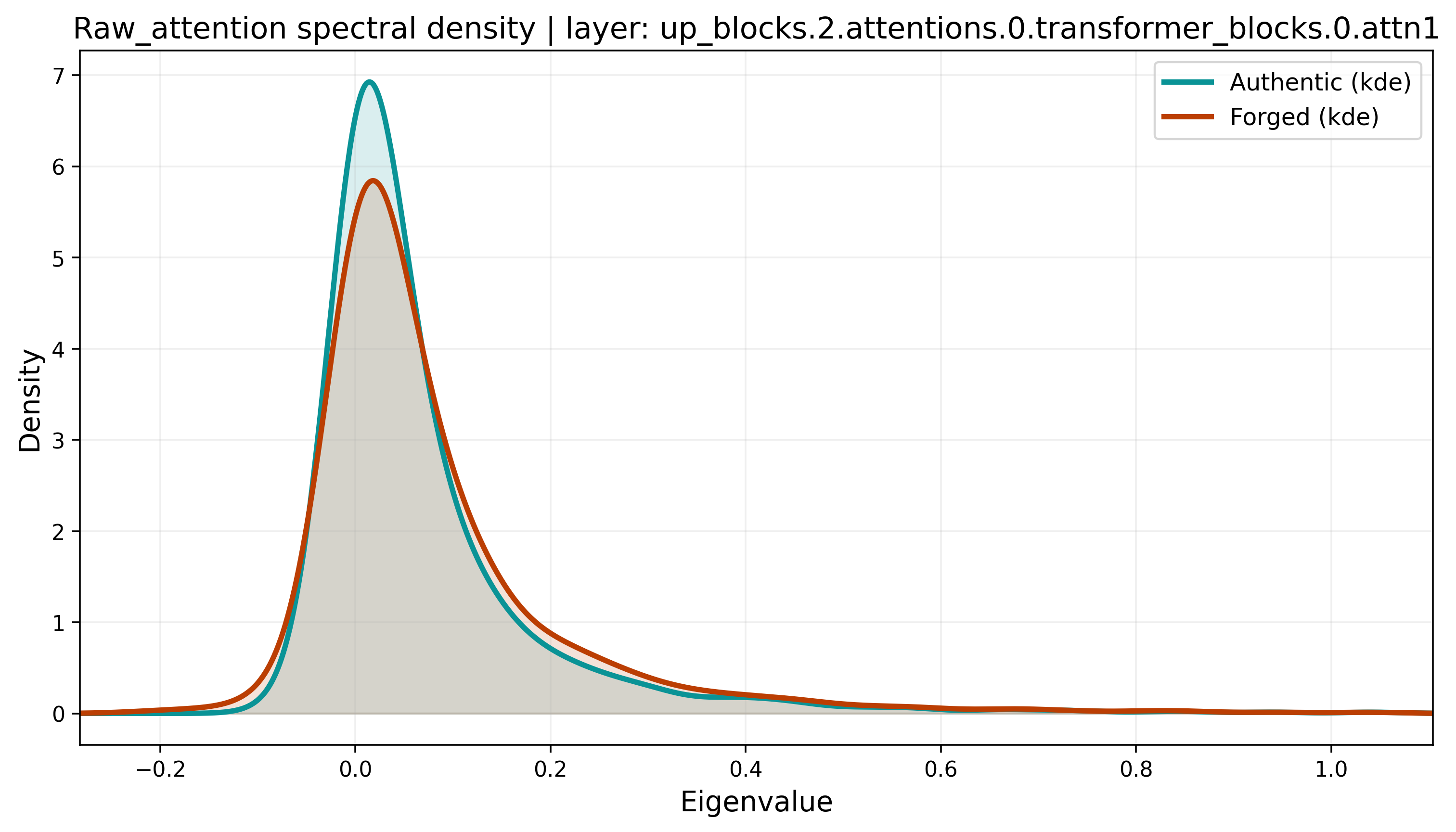}
    \caption{Raw attention spectral density}
\end{subfigure}
\hfill
\begin{subfigure}[t]{0.48\linewidth}
    \centering
    \includegraphics[width=\linewidth]{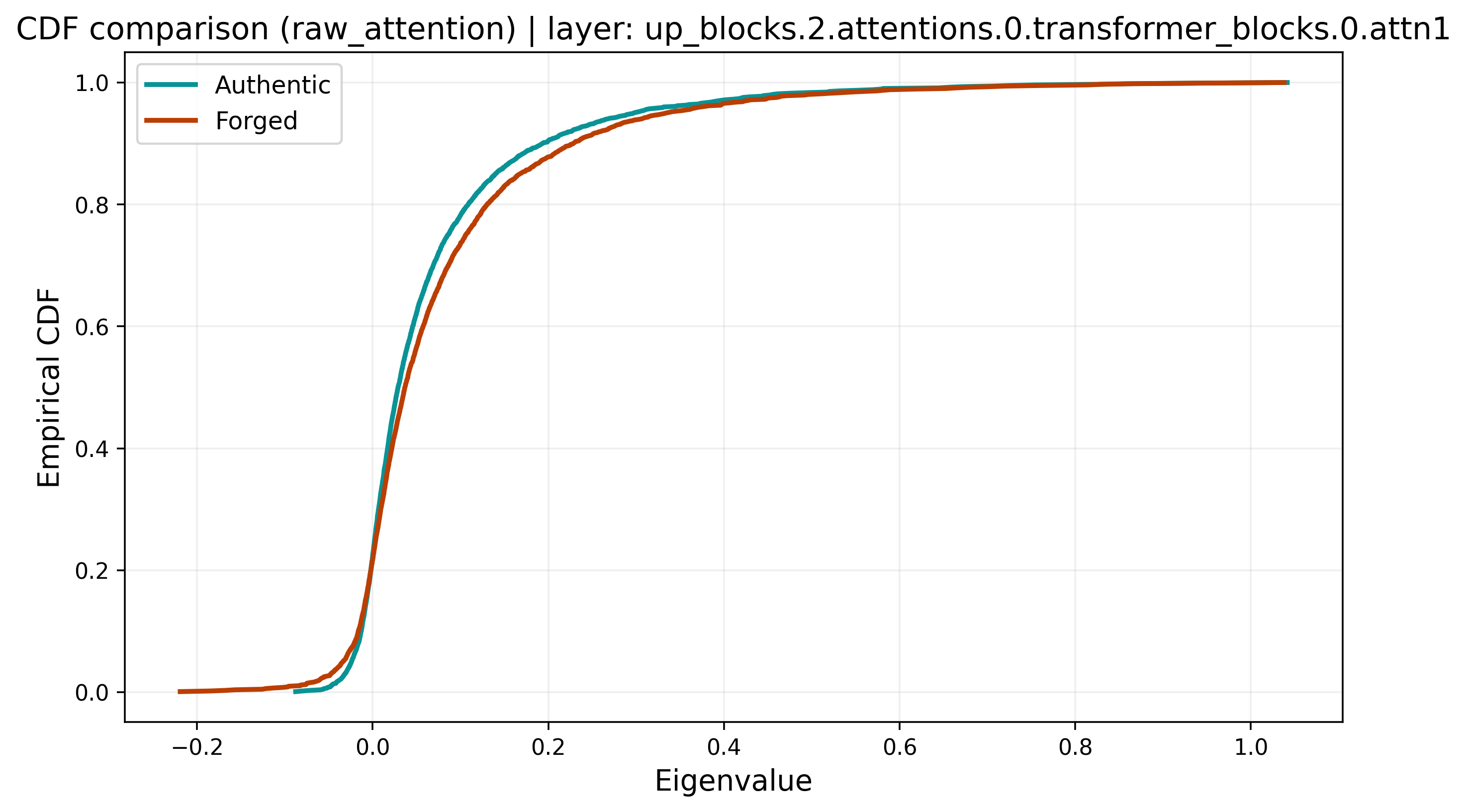}
    \caption{Raw attention CDF}
\end{subfigure}
\caption{\textbf{Raw-attention per-image comparison on pair \texttt{013}.} At the same layer, the raw attention ESDs of authentic and forged images differ mainly near $\sigma = 0$ and in the heavy right tail; the Laplacian representation (\Cref{fig:supp_pairwise_laplacian}) rebalances this so that the informative shift is concentrated near $\lambda = 1$, where the duplication-band hypothesis predicts it should lie.}
\label{fig:supp_pairwise_raw}
\end{figure}

\paragraph{Interpretation.}
The pair-level figures illustrate at single-image resolution the same mechanism the main paper quantifies in aggregate: a real copy-move operation is visually imperceptible and cannot be separated reliably by any single pixel-level statistic, yet its \emph{graph-spectral fingerprint}---measured as a small but consistent shift in the Laplacian ESD---is already visible at the level of one pair.
This is the single-image analogue of Proposition~1 of the main paper: the approximate subgraph duplication perturbs the normalized Laplacian by a small $\Delta$, producing a matched small shift in the ESD, whose Wasserstein magnitude is the basic scalar used by our detector.

\FloatBarrier
\section{Extended External-Dataset Ablations}
\label{supp:external}

This section reports the full archived feature-group ablations for the three external copy-move datasets summarised in the main paper: MICC-F220, CoMoFoD, and COVERAGE.
For each dataset, all configurations reuse the same training-free Stable Diffusion v1.5 pipeline; only the authentic reference statistics (median/MAD per layer per feature) are recomputed on the dataset itself.
No forgery-specific training is performed on any dataset.
Columns: \emph{Spec.} --- spectrum type; \emph{Bundle} --- feature group; \emph{Scale} --- robust ($z$ with median/MAD) or plain ($z$ with mean/std); \emph{Fusion} --- unweighted mean of top-$k$ per-layer scores vs.\ softmax-weighted fusion with AUROC or reliability scores; \emph{$k$} --- number of fused layers.
All CIs are bootstrap 95\% intervals.
Rows are truncated to the Laplacian family (which dominates in almost all settings); the raw-spectrum entries are available in the source CSVs but never better than the best Laplacian configuration by more than $0.02$ AUROC on any of the three datasets.

\begin{table}[!htbp]
\centering
\caption{\textbf{MICC-F220 Laplacian ablation grid.} $n = 44$. Best row in \textbf{bold}.}
\label{tab:supp_micc_ablation}
\scriptsize
\setlength{\tabcolsep}{3.2pt}
\begin{tabular}{@{}llllrcccc@{}}
\toprule
\textbf{Bundle} & \textbf{Scale} & \textbf{Fusion} & $\boldsymbol{k}$ & \textbf{AUROC} & \textbf{CI} & \textbf{AUPRC} & \textbf{TPR@1\%} & \textbf{TPR@5\%} \\
\midrule
$W_1$-only          & robust & unweighted           & 3 & 0.589 & [0.40, 0.75] & 0.665 & 0.23 & 0.23 \\
$W_1$-only          & robust & softmax-AUROC        & 3 & 0.566 & [0.38, 0.72] & 0.654 & 0.23 & 0.23 \\
$W_1$-only          & robust & softmax-reliab.      & 3 & 0.554 & [0.37, 0.72] & 0.646 & 0.23 & 0.23 \\
Transport           & robust & unweighted           & 3 & 0.632 & [0.45, 0.80] & 0.670 & 0.05 & 0.23 \\
Transport           & robust & softmax-AUROC        & 3 & 0.616 & [0.44, 0.80] & 0.633 & 0.00 & 0.23 \\
Transport           & robust & softmax-reliab.      & 3 & 0.599 & [0.42, 0.78] & 0.619 & 0.00 & 0.23 \\
Filter-bank         & robust & unweighted           & 3 & 0.616 & [0.45, 0.79] & 0.685 & 0.27 & 0.27 \\
Filter-bank         & robust & softmax-AUROC        & 3 & 0.641 & [0.47, 0.80] & 0.691 & 0.18 & 0.23 \\
Filter-bank         & robust & softmax-reliab.      & 3 & 0.638 & [0.46, 0.80] & 0.686 & 0.18 & 0.18 \\
Transport+filter    & robust & unweighted           & 3 & 0.686 & [0.50, 0.84] & 0.737 & 0.32 & 0.32 \\
Transport+filter    & robust & softmax-AUROC        & 3 & 0.705 & [0.52, 0.85] & 0.736 & 0.23 & 0.27 \\
Transport+filter    & robust & softmax-reliab.      & 3 & 0.700 & [0.50, 0.84] & 0.727 & 0.23 & 0.23 \\
Transport+dup.      & robust & unweighted           & 3 & 0.657 & [0.46, 0.82] & 0.643 & 0.00 & 0.09 \\
All features        & robust & unweighted           & 3 & 0.663 & [0.49, 0.82] & 0.698 & 0.05 & 0.23 \\
All features        & robust & softmax-AUROC        & 3 & 0.709 & [0.54, 0.87] & 0.759 & 0.09 & 0.45 \\
All features        & robust & softmax-reliab.      & 3 & 0.690 & [0.51, 0.86] & 0.755 & 0.18 & 0.36 \\
All + controls      & robust & unweighted           & 3 & 0.688 & [0.51, 0.86] & 0.736 & 0.14 & 0.14 \\
All + controls      & robust & softmax-AUROC        & 3 & 0.696 & [0.52, 0.87] & 0.759 & 0.23 & 0.32 \\
All + controls      & robust & softmax-reliab.      & 3 & 0.696 & [0.53, 0.85] & 0.748 & 0.27 & 0.32 \\
\textbf{All features} & \textbf{robust} & \textbf{softmax-AUROC} & \textbf{2} & \textbf{0.752} & \textbf{[0.59, 0.90]} & \textbf{0.721} & \textbf{0.09} & \textbf{0.23} \\
\bottomrule
\end{tabular}
\end{table}

\begin{table}[!htbp]
\centering
\caption{\textbf{CoMoFoD Laplacian ablation grid (best archived row).} $n = 80$. The best archived CoMoFoD configuration uses the plain $z$-score scaling and unweighted fusion with $k = 5$; higher $k$ values also pick up the decoder-heavy mid-layers that dominate this dataset's spectral signal.}
\label{tab:supp_comofod_ablation}
\small
\setlength{\tabcolsep}{4pt}
\begin{tabular}{@{}lcccc@{}}
\toprule
\textbf{Configuration} & \textbf{AUROC} & \textbf{CI} & \textbf{AUPRC} & \textbf{TPR@5\%} \\
\midrule
Lap., $W_1$-only, plain, $k{=}5$, unweighted (\textbf{best}) & \textbf{0.774} & \textbf{[0.651, 0.886]} & \textbf{0.833} & \textbf{0.550} \\
Lap., $W_1$-only, robust, $k{=}3$, unweighted & 0.741 & [0.613, 0.857] & 0.801 & 0.488 \\
Lap., transport, plain, $k{=}5$, unweighted  & 0.732 & [0.601, 0.847] & 0.782 & 0.475 \\
Lap., all features, plain, $k{=}5$, unweighted & 0.719 & [0.589, 0.833] & 0.760 & 0.463 \\
Lap., transport+filter, plain, $k{=}5$, softmax-AUROC & 0.709 & [0.577, 0.824] & 0.751 & 0.438 \\
Raw, $W_1$-only, plain, $k{=}5$, unweighted  & 0.685 & [0.549, 0.803] & 0.707 & 0.400 \\
\bottomrule
\end{tabular}
\end{table}

\begin{table}[!htbp]
\centering
\caption{\textbf{COVERAGE Laplacian ablation grid (best archived row).} $n = 40$. The smaller sample inflates CI width substantially; the best setting retains Laplacian features and softmax-reliability fusion with $k = 2$.}
\label{tab:supp_coverage_ablation}
\small
\setlength{\tabcolsep}{4pt}
\begin{tabular}{@{}lcccc@{}}
\toprule
\textbf{Configuration} & \textbf{AUROC} & \textbf{CI} & \textbf{AUPRC} & \textbf{TPR@5\%} \\
\midrule
Lap., $W_1$-only, robust, $k{=}2$, softmax-reliab.\ (\textbf{best}) & \textbf{0.673} & \textbf{[0.504, 0.828]} & \textbf{0.653} & \textbf{0.100} \\
Lap., $W_1$-only, robust, $k{=}3$, unweighted & 0.655 & [0.488, 0.814] & 0.636 & 0.125 \\
Lap., transport, plain, $k{=}3$, unweighted  & 0.640 & [0.472, 0.804] & 0.621 & 0.150 \\
Lap., transport+dup., robust, $k{=}3$, softmax-AUROC & 0.635 & [0.469, 0.797] & 0.620 & 0.100 \\
Raw, $W_1$-only, plain, $k{=}3$, unweighted  & 0.638 & [0.472, 0.802] & 0.609 & 0.075 \\
\bottomrule
\end{tabular}
\end{table}

\paragraph{Dataset-level observations.}
Three patterns emerge consistently across the three external grids:
\emph{(i)} The Laplacian wins the matched all-plus-controls comparison on MICC-F220 and COVERAGE and loses marginally on CoMoFoD; but the best-archived row on each dataset uses Laplacian features.
\emph{(ii)} MICC-F220 prefers richer bundles (all features $+$ controls), CoMoFoD peaks on the minimalist $W_1$-only setting, and COVERAGE sits between the two.
\emph{(iii)} The fusion depth $k$ that is best on the main RecodAI-LUC benchmark ($k = 5$) is \emph{not} optimal on the external sets: MICC-F220 peaks at $k = 2$ and COVERAGE at $k = 2$.
This is consistent with the main-paper observation that the signal is concentrated in a few decoder layers, and smaller external sets give unstable estimates for layers further down the ranking.

\FloatBarrier
\section{External-Dataset Per-Layer Wasserstein Rankings}
\label{supp:external_layers}

For completeness, \Cref{tab:supp_external_layers} reports the full 16-layer Wasserstein rankings for each external dataset.
Layers are sorted by pooled full-spectrum Wasserstein $W_1^F$ within each column.
The ranking is dominated by decoder layers (\texttt{up\_blocks}) on all three datasets, reproducing the main-benchmark pattern.

\begin{table}[!htbp]
\centering
\caption{\textbf{Per-layer pooled Wasserstein distances on the three external datasets.} $W_1^F$: full ($\times 10^{-3}$). $W_1^T$: tail ($\times 10^{-3}$). $d$: Cohen's $d$. Image-level AUROC uses per-image Wasserstein at that layer against the dataset's authentic reference. Cells sorted within each dataset by $W_1^F$.}
\label{tab:supp_external_layers}
\scriptsize
\setlength{\tabcolsep}{3pt}
\begin{tabular}{@{}l|rrrc|rrrc|rrrc@{}}
\toprule
& \multicolumn{4}{c|}{\textbf{MICC-F220}} & \multicolumn{4}{c|}{\textbf{CoMoFoD}} & \multicolumn{4}{c}{\textbf{COVERAGE}} \\
\textbf{Layer} & $W_1^F$ & $W_1^T$ & $d$ & AUROC & $W_1^F$ & $W_1^T$ & $d$ & AUROC & $W_1^F$ & $W_1^T$ & $d$ & AUROC \\
\midrule
\texttt{up3.attn0}   & 1.62 & 4.26 & 0.05 & 0.45 & 8.76 & 33.6 & 0.25 & 0.45 & 12.70 & 46.3 & 0.33 & 0.59 \\
\texttt{up3.attn1}   & 3.34 & 8.97 & 0.06 & 0.49 & 5.81 & 15.0 & 0.21 & 0.44 & 8.50 & 24.3 & 0.28 & 0.59 \\
\texttt{up2.attn0}   & 1.22 & 1.86 & -0.06 & 0.41 & 5.61 & 12.4 & 0.10 & 0.46 & 6.73 & 12.0 & 0.09 & 0.55 \\
\texttt{up2.attn1}   & 1.89 & 2.01 & 0.08 & 0.48 & 4.35 & 9.15 & 0.12 & 0.45 & 6.89 & 12.6 & 0.17 & 0.56 \\
\texttt{up3.attn2}   & 2.35 & 9.72 & 0.11 & 0.43 & 4.28 & 12.8 & 0.16 & 0.45 & 6.93 & 17.5 & 0.29 & 0.56 \\
\texttt{up1.attn1}   & 1.96 & 8.95 & -0.08 & 0.37 & 4.37 & 4.36 & 0.09 & 0.47 & 3.98 & 1.53 & 0.03 & 0.48 \\
\texttt{up1.attn0}   & 1.12 & 6.74 & 0.06 & 0.49 & 1.66 & 3.78 & 0.04 & 0.46 & 1.36 & 2.09 & 0.01 & 0.50 \\
\texttt{up1.attn2}   & 1.02 & 4.94 & 0.02 & 0.46 & 0.91 & 2.87 & 0.07 & 0.50 & 0.35 & 1.52 & 0.03 & 0.49 \\
\texttt{up2.attn2}   & 0.80 & 5.23 & 0.12 & 0.47 & 0.99 & 3.01 & 0.09 & 0.47 & 1.16 & 4.30 & 0.11 & 0.52 \\
\texttt{mid.attn0}   & 3.18 & 9.84 & 0.19 & 0.49 & 1.26 & 5.14 & 0.01 & 0.48 & 0.57 & 2.43 & -0.02 & 0.49 \\
\texttt{down2.attn1} & 1.82 & 4.66 & -0.09 & 0.50 & 1.60 & 2.40 & 0.04 & 0.47 & 2.09 & 1.86 & 0.07 & 0.51 \\
\texttt{down2.attn0} & 0.94 & 3.12 & -0.06 & 0.49 & 1.29 & 2.35 & 0.00 & 0.48 & 0.97 & 1.33 & -0.02 & 0.51 \\
\texttt{down1.attn1} & 0.98 & 3.18 & 0.03 & 0.50 & 1.63 & 4.67 & 0.00 & 0.51 & 2.74 & 5.31 & 0.00 & 0.58 \\
\texttt{down1.attn0} & 0.43 & 1.61 & 0.03 & 0.48 & 0.37 & 0.41 & -0.02 & 0.49 & 0.86 & 1.02 & -0.04 & 0.50 \\
\texttt{down0.attn1} & 0.37 & 1.28 & 0.02 & 0.49 & 0.23 & 0.20 & -0.01 & 0.49 & 0.94 & 2.08 & -0.02 & 0.49 \\
\texttt{down0.attn0} & 0.17 & 0.72 & 0.02 & 0.49 & 0.20 & 0.93 & -0.02 & 0.49 & 0.21 & 0.62 & -0.01 & 0.49 \\
\bottomrule
\end{tabular}
\end{table}

\paragraph{Takeaways.}
On all three datasets the decoder layer \texttt{up\_blocks.3.attentions.0} is among the top three by pooled $W_1^F$, reproducing the RecodAI-LUC pattern.
COVERAGE additionally places \texttt{up\_blocks.3.attn1} and the mid-resolution \texttt{up\_blocks.2.*} family near the top, explaining why softmax-reliability fusion with small $k$ works best there.
CoMoFoD shows the cleanest monotonic structure ($W_1^F$ decreasing from $8.76$ at \texttt{up3.attn0} down to $0.20$ at \texttt{down0.attn0}), which is also why the simplest $W_1$-only detector suffices on that dataset.
The small absolute Cohen's $d$ values on MICC-F220 reflect its small sample size ($n_{\text{auth}} = n_{\text{forg}} = 110$ per layer) rather than a weaker spectral shift per se; AUROC values are above chance at the top-ranked decoder layers on all three datasets.

\FloatBarrier
\section{Notes on Reproducibility}
\label{supp:repro}

All external ablation CSVs and per-layer metric CSVs used to populate \Cref{tab:supp_micc_ablation,tab:supp_comofod_ablation,tab:supp_coverage_ablation,tab:supp_external_layers} are archived in the \texttt{Experiments\_for\_other\_datasets/} directory of the accompanying code release, in the \texttt{layer\_distributed\_wasserstein/} subdirectories of each dataset folder.
The pairwise case-study figures in \Cref{supp:pairwise} are generated from the authentic and forged Laplacian eigenvalue files \texttt{authentic\_up\_blocks\_2\_attn0\_lap\_eigs.csv} and \texttt{forged\_up\_blocks\_2\_attn0\_lap\_eigs.csv} stored alongside \texttt{pairwise\_summary.csv} in the \texttt{Pairwise analysis/0013\_013/} directory; these files make the case study fully reproducible from raw eigenvalues.